\documentclass[lettersize,journal]{IEEEtran}
\usepackage{amsmath,amsfonts}
\usepackage{array}
\usepackage{textcomp}
\usepackage{stfloats}
\usepackage{url}
\usepackage{verbatim}
\usepackage{graphicx}
\usepackage{subcaption}
\usepackage[export]{adjustbox}
\usepackage[ruled,vlined]{algorithm2e}
\usepackage{multirow}
\usepackage[noend]{algpseudocode}
\usepackage{subcaption}
\usepackage{xcolor}
\usepackage{hyperref}
\usepackage{comment}
\usepackage{multicol}
\usepackage{xcolor}
\usepackage{float}
\usepackage{tikz}
\usepackage{booktabs} 
\usepackage{rotating}
\usepackage{pdflscape}  

\def\BibTeX{{\rm B\kern-.05em{\sc i\kern-.025em b}\kern-.08em
		T\kern-.1667em\lower.7ex\hbox{E}\kern-.125emX}}
\usepackage{balance}

\begin{document}

	\title{Enhancing MedSAM with a Lightweight Box Predictor for Medical Image Segmentation}
	
	\author{
    Amirhossein Movahedisefat\textsuperscript{1},
    Amirreza Fateh\textsuperscript{1}, Mohammad Reza Mohammadi\textsuperscript{1,*} \\
    \textsuperscript{1} School of Computer Engineering, Iran University of Science 
    and Technology (IUST), Tehran, Iran \\
    \textsuperscript{*} Corresponding author. mrmohammadi@iust.ac.ir

}
	
	\maketitle

	\begin{abstract}\textcolor{black}{
Semantic segmentation in medical imaging is a critical yet challenging task due to data scarcity and high variability across modalities. While foundation models like the Segment Anything Model (SAM) show promise, they often struggle with medical images without specific adaptation. Moreover, point prompts, despite being the most natural form of user interaction, provide insufficient spatial context for reliable segmentation, particularly when target structures are irregular or poorly contrasted. In this paper, we propose an enhanced segmentation framework that integrates a lightweight Box Predictor module into the MedSAM architecture. The Box Predictor estimates an approximate bounding box from a single user click using localized image embedding features, providing spatial guidance that reduces the ambiguity of point prompts, while introducing only 1.6M additional parameters and negligible inference overhead. We introduce a two-stage training pipeline where the Box Predictor is trained independently before being integrated into MedSAM. To validate the generalization capability of our method, we conduct extensive evaluations on four diverse datasets (FLARE22, BRISC, BUSI, LungSegDB) spanning distinct imaging modalities, including CT, MRI, and Ultrasound. Our method improves segmentation accuracy and robustness across varied anatomical structures and imaging domains, achieving Dice scores of 0.89 (BUSI), 0.93 (FLARE22), 0.88 (BRISC), and 0.98 (LungSegDB). Code is available at \url{https://github.com/Amirhosseinmovahedi/MedSAM-BoxPredictor}}
\end{abstract}

	\begin{IEEEkeywords}
		Medical Image Segmentation, Segment Anything Model, MedSAM, Box Predictor, Deep Learning.
	\end{IEEEkeywords}
	
	\section{Introduction}
	\label{sec:intro}
	Semantic segmentation is a fundamental task in computer vision that assigns a class label to every pixel in an image, effectively distinguishing the various structures present \cite{yuan2026icad,fateh2026adapting,fateh2025msdnet}. In medical imaging, this task is important for applications such as identifying organs, marking abnormal areas like tumors, and analyzing anatomical structures \cite{yang2025cdsg,ying2025sam}. Accurate segmentation supports clinical decision-making and assists in planning treatments or surgeries. However, medical image segmentation is challenging due to the low contrast of medical images and their wide variation across modalities (e.g., CT, MRI, ultrasound), and differ significantly between patients \cite{24}. Modern deep learning-based segmentation methods require large amounts of labeled data to generalize well across such diverse conditions. While natural image segmentation benefits from large-scale benchmarks such as COCO \cite{shi2025sit}, creating comparable annotated datasets in medical imaging is significantly more challenging and expensive, requiring expert clinicians to manually delineate structures in each image. These factors make it difficult to train segmentation models that perform well across diverse medical tasks.
	
	Deep learning–based architectures, such as U-Net \cite{19} and DeepLabV3+ \cite{18}, have substantially improved the accuracy and robustness of segmentation systems across a variety of imaging domains. Nevertheless, their success largely depends on access to large-scale annotated datasets, which are rarely available in medical contexts \cite{32,33}. This challenge has motivated a shift toward pre-trained models that can transfer knowledge from large datasets to specific clinical applications. These models offer strong generalization capabilities and can be fine-tuned for specific medical imaging applications, effectively reducing the reliance on large annotated datasets and improving performance in settings with limited labeled data \cite{34,35}. By leveraging rich pre-trained representations, foundation models provide a powerful alternative to training deep networks from scratch.
	
	Among recent foundation models, the Segment Anything Model (SAM) \cite{2} has emerged as a general-purpose segmentation framework capable of producing accurate masks from simple user prompts such as points, boxes, or text. However, applying SAM directly to medical images is non-trivial due to the significant visual and structural differences between natural and medical imagery \cite{25,26,27,28}. To address this, MedSAM \cite{3} was introduced as an adaptation of SAM specifically fine-tuned for medical datasets. While MedSAM improves over the original SAM by training on diverse medical imaging data, the heterogeneity across imaging modalities, anatomical regions, and acquisition protocols means that further domain-specific fine-tuning is often necessary to achieve optimal performance on particular clinical tasks. However, such fine-tuning introduces additional challenges. First, the model’s vast number of parameters makes full fine-tuning computationally infeasible for most medical research settings. Second, SAM-based models are highly sensitive to the quality of input prompts, with this issue becoming particularly severe in medical imaging where regions of interest are often small, irregular, or poorly contrasted \cite{36,37}. Improving prompt robustness is thus critical for reliable medical image segmentation.
	
	To mitigate the challenges associated with prompt dependency and limited computational resources, we propose an enhanced segmentation framework that integrates a lightweight Box Predictor module with the MedSAM architecture. Our approach is motivated by a specific observation: while a single point prompt may be convenient for users, it often provides insufficient spatial context for accurate segmentation, particularly when the target region has ambiguous boundaries or lies in low-contrast areas. To overcome this issue, we introduce a Box Predictor module that estimates a coarse but informative bounding box from minimal input (e.g., a single point), which then serves as an additional spatial prior to guide the prompt encoder and mask decoder toward more accurate region localization. Importantly, the Box Predictor is trained separately with low computational cost and subsequently integrated into the MedSAM pipeline, allowing the model to leverage both the learned medical representations of MedSAM and the geometric guidance of the predicted box. \textcolor{black}{This integration improves the robustness of the segmentation process while maintaining efficiency, with the primary focus on enhancing the practical usability of MedSAM in realistic clinical workflows rather than introducing a fundamentally new segmentation paradigm.}
	
	The main contributions of this study are summarized as follows:
    \begin{itemize}
       \item \textbf{A lightweight Box Predictor module:} We introduce a lightweight Box Predictor that converts a single point prompt into an approximate bounding box, providing spatial priors that stabilize MedSAM when point prompts alone lead to poor performance, especially under inaccurate or poorly placed prompts.
        \item \color{black}\textbf{A two-stage training pipeline:} The Box Predictor is first trained independently, then integrated as fixed spatial guidance while MedSAM's Prompt Encoder and Mask Decoder are optionally fine-tuned. This design enables robust segmentation even when MedSAM remains frozen, suitable for resource-constrained settings.\color{black}
        \item \textbf{Systematic evaluation of prompt configurations and domain generalization:} We conduct a comprehensive assessment of six MedSAM configurations, methodically examining the interaction between prompt types (point, box, and hybrid) and model states (frozen versus fine-tuned). This analysis extends across four heterogeneous medical imaging datasets encompassing MRI, CT, and Ultrasound. Our goal is to demonstrate the framework's consistent robustness across diverse anatomical regions and imaging modalities.
        \item \color{black}\textbf{Impact of Prompt Variations on Segmentation Accuracy:} We investigate how modifications to input points and predicted bounding boxes influence segmentation accuracy, offering guidance for effective prompt design.\color{black}
    \end{itemize}

	\section{Related Work}
	
	\subsection{Semantic Segmentation}
	Semantic segmentation has long been a central task in computer vision, forming the foundation for a variety of applications across medical and non-medical domains. Classical architectures such as U-Net \cite{19}, DeepLabV3+ \cite{18}, and their later variants significantly improved segmentation performance by leveraging encoder–decoder structures. In medical imaging, these models have been widely adopted due to their ability to learn spatially precise representations even from limited supervision. Nevertheless, their effectiveness is constrained by the limited availability of large, high-quality annotated medical datasets and by the substantial variability across imaging modalities such as CT, MRI, and ultrasound \cite{32,33}. These limitations have motivated a shift toward large pre-trained vision models that can transfer learned representations to downstream medical tasks with minimal domain-specific labels \cite{34,35}.
	
	\subsection{SAM-Based Models}
	The introduction of the Segment Anything Model (SAM) \cite{2} marked a major shift toward foundation models capable of general-purpose segmentation. SAM incorporates a ViT-based image encoder, a prompt encoder that processes interactive inputs such as points or boxes, and a mask decoder that synthesizes segmentation outputs from these multimodal representations. Although SAM performs well on natural images, its performance degrades when applied directly to medical images due to domain gaps such as low contrast, modality-specific texture patterns, and diverse anatomical structures.
	
	To mitigate this issue, Medical SAM (MedSAM) \cite{3} fine-tunes SAM on large collections of medical images while preserving its architectural design. As shown in \cite{3}, the model retains the original encoders and decoder but updates their parameters using medical data to achieve better adaptation to clinical imaging characteristics. This strategy improves robustness to the unique intensity distributions and noise profiles present in modalities such as CT and MRI, demonstrating that domain-aligned pre-training is essential for high-quality medical segmentation.
	
	\textcolor{black}{Building on SAM, SAM2 \cite{40} extended the framework to video and volumetric data by introducing a memory mechanism that propagates segmentation across frames. Although originally designed for sequential and 3D inputs, SAM2 can be applied to 2D medical images by treating each slice independently, albeit without leveraging its temporal modeling capabilities. MedSAM2 \cite{39} adapts this architecture to the medical domain through domain-specific fine-tuning, similarly applied here in a frame-by-frame 2D inference mode. SAM2UNet \cite{38}, in contrast, replaces the interactive prompting paradigm with a fully automatic segmentation framework that leverages the hierarchical Hiera backbone of SAM2 within a U-shaped encoder–decoder architecture, coupled with lightweight adapters for efficient fine-tuning, eliminating the need for user-provided spatial prompts during inference.} \textcolor{black}{Most recently, SAM 3~\cite{44} introduced a unified model supporting both Promptable Visual Segmentation (PVS) and Promptable Concept Segmentation (PCS). PVS uses points, bounding boxes, or masks to specify an individual target object, whereas PCS uses a short noun phrase, image exemplars, or both to detect and segment all instances matching a visual concept. A controlled zero-shot study on 3D medical data~\cite{47} compared SAM 2 and SAM 3 in the PVS setting using matched single-click, multi-click, bounding-box, and mask prompts.}
	
	A more comprehensive investigation is provided by \cite{31}, which systematically compares multiple SAM fine-tuning strategies across interactive 2D/3D and semantic segmentation tasks, showing that domain-specific fine-tuning significantly enhances interactive performance while offering benefits for downstream semantic segmentation.
	
	Several works further analyze or extend SAM by modifying or augmenting the prompt encoder. Because SAM’s behavior is heavily guided by user-provided prompts, research has focused on improving the model’s ability to interpret imperfect, imprecise, or ambiguous inputs. For example, the robust fine-tuning strategy proposed in \cite{7} introduces random perturbations to bounding-box prompts during training, enabling the model to better tolerate inaccuracies during inference. By exposing SAM to a continuum of perturbed prompts, the method enhances generalization in clinical workflows where user interaction may not be perfectly precise.
	
	\color{black}
	\subsection{Prompt Enhancement and Generation}
	
	A separate line of work investigates reducing SAM’s dependence on manually supplied prompts by generating spatial cues directly from the image. The framework proposed in \cite{6} introduces a module that produces candidate masks and bounding boxes automatically using multilevel encoder features. Different configurations of mask and box generation strategies are evaluated, demonstrating that combining learned and extracted bounding boxes yields the most reliable prompt set. Extending this automated generation paradigm to cross-domain scenarios, \cite{36} introduces uncertainty modeling of shallow features to generate bounding box prompts that remain robust under domain shift between training and test distributions. This direction aims to support fully automated segmentation pipelines where human interaction is either impractical or undesirable.
	
	Another research thread focuses on improving the quality of prompts rather than generating them. Learning-based prompt refinement methods, such as those in \cite{10,11,13}, develop specialized modules to derive more accurate bounding boxes, point prompts, or embeddings from initial coarse inputs. For instance, \cite{10} optimizes spatial and semantic prompts in the embedding space to address limited search constraints, while \cite{11} refines low-quality box prompts by predicting offsets through a prompt refinement module and enhances attention on relevant regions using self-information features, achieving robust segmentation even under misaligned or noisy prompts. \cite{13} employs a prompt adapter to refine sparse and dense representations, improving SAM's sensitivity to fine details and uncertain areas without altering the core model. In contrast to these refinement-based approaches, \cite{37} proposes a training-free method that generates point prompts from a single reference annotation using evidential learning and spatial ensemble of perturbations, eliminating the need for manual prompts or large labeled datasets.
	
	Beyond spatial cues, several works explore integrating vision–language models (VLMs) to provide semantic prompts for domain-specific tasks. While SAM does not natively support text-based prompting, coupling it with CLIP \cite{9} allows the model to interpret basic textual descriptions within segmentation workflows. However, because standard CLIP is trained on natural images, it often struggles with medical terminology and modality-specific visual patterns. To address this limitation, domain-adapted variants such as MedCLIP \cite{8} have been introduced to more effectively encode medical text–image relationships. These models enable SAM-based systems to incorporate semantic information related to anatomical structures, thereby expanding the types of prompts that can guide segmentation in clinical settings. Building on this idea, the multimodal prompting strategy in \cite{30} integrates CLIP-derived saliency cues, VQA-generated anatomical hints, and GPT-based textual descriptions to provide richer semantic prompts that reduce SAM’s reliance on precise human inputs. Complementing these approaches, \cite{29} uses large language models to decompose radiology reports into organ-level textual descriptions and aligns them with 3D CT volumes through multi-granularity contrastive learning, enabling fully text-driven CT segmentation without any spatial prompts.\color{black} \textcolor{black}{
Unlike prior prompt refinement methods that primarily optimize or iteratively adjust existing prompts through additional attention or refinement mechanisms, our approach focuses on learning lightweight spatial priors directly from localized MedSAM embedding features using minimal user interaction. The proposed Box Predictor operates as a computationally efficient auxiliary module that can be integrated into MedSAM without modifying its core architecture, making it suitable for resource-constrained medical imaging scenarios.
}

	\section{Proposed Method}
	\label{sec:method}
	
	\subsection{Overview of the Proposed Method}
	In this section, the proposed method for enhancing the segmentation performance of the MedSAM model in medical image analysis is presented. Considering the significant impact of prompt design in interactive segmentation frameworks, this work focuses on developing an automatic prompt generation mechanism that leverages image-derived features. The objective is to provide the model with richer and more spatially informative prompts. To this end, a dedicated auxiliary module, referred to as the Box Predictor, is designed to generate a coarse spatial prior for the target region. Given localized embeddings extracted from the Image Encoder of MedSAM, the Box Predictor estimates an approximate bounding box around the region of interest (ROI).
	
\begin{figure*}[t]
		\centering
		
		\centering
		\includegraphics[width=0.9\linewidth]{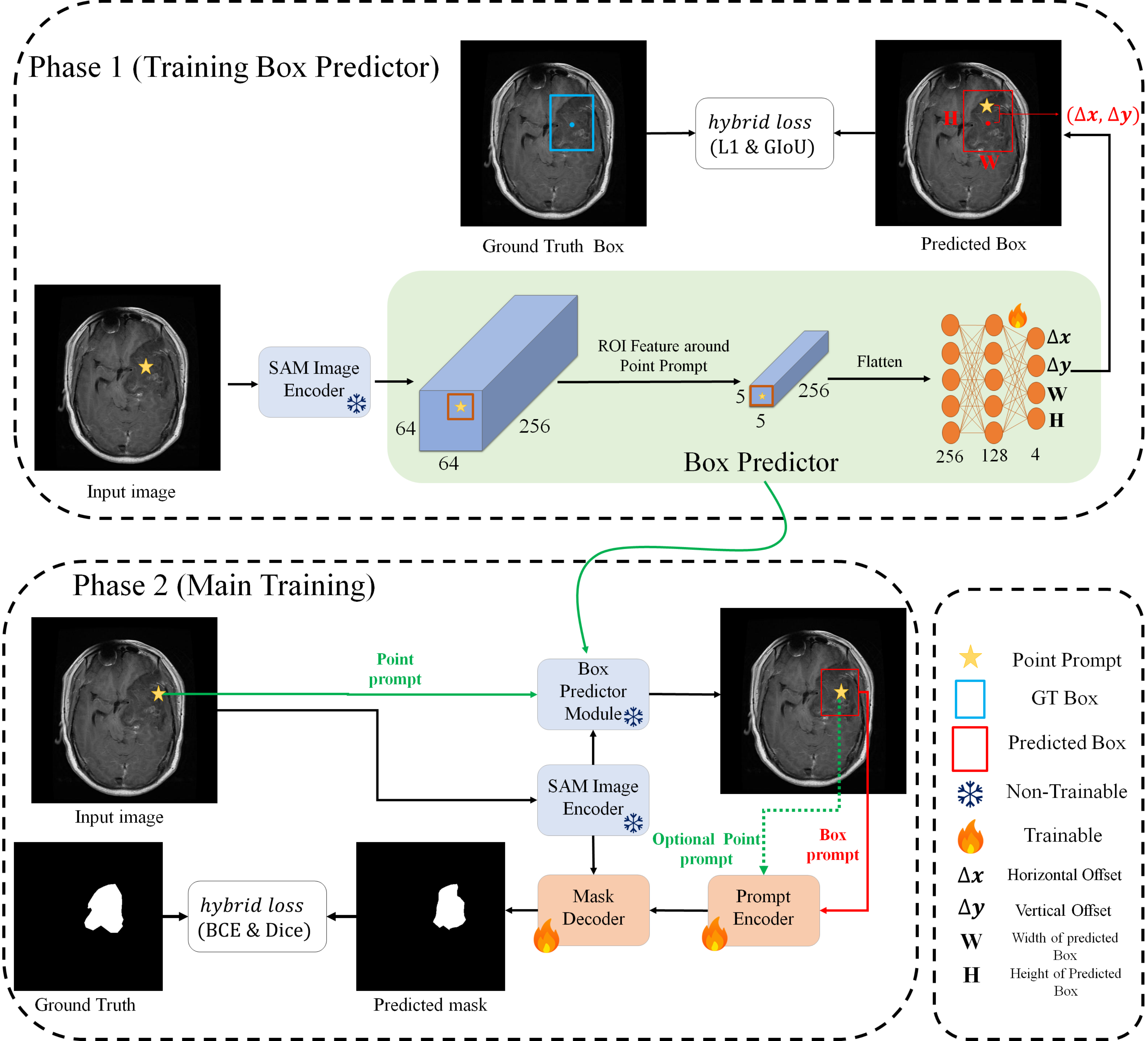}
		\caption{Schematic illustration of the two-stage training framework. In Stage 1, a 5$\times$5 local patch from the 64$\times$64$\times$256 image embedding around the point prompt is flattened and passed to the Box Predictor (MLP), which outputs bounding box parameters (W,H,$\Delta$x,$\Delta$y). The model is trained using a combined loss function $L=L_{L1}+L_{GIoU}$, where the predicted box (red) is compared against the ground truth box (blue). In Stage 2, the trained Box Predictor provides the predicted box prompt, which is fed into the Prompt Encoder. The box can be used alone or in combination with point prompts to guide segmentation. During this stage, the Image Encoder remains frozen, while the Prompt Encoder and Mask Decoder are trainable. The final output is the predicted segmentation mask, compared against the ground truth mask for segmentation loss computation.}
		\label{fig:framework}
	\end{figure*}

	During the integration phase, the pre-trained Box Predictor and Image Encoder are kept frozen to provide robust input features, while the Prompt Encoder and Mask Decoder are either kept frozen or fine-tuned to adapt the model to the medical imaging domain. In the frozen setup, the Box Predictor is crucial for correcting low-quality prompts and ensuring reliable segmentation. When the Prompt Encoder and Mask Decoder are fine-tuned, the model can better specialize to the characteristics of the specific dataset, further improving accuracy and robustness. This selective training strategy balances the preservation of general visual knowledge from the foundation model with efficient domain-specific adaptation using limited data and computational resources. The overall architecture of the proposed method, including the interaction between the Box Predictor and the MedSAM framework, is illustrated in Figure~\ref{fig:framework}.

	\subsection{Architecture of the Box Predictor}
	The Box Predictor module is designed to generate an approximate bounding box based on high-level visual features extracted from the Image Encoder and a single input point. The input point acts as a simulated user click and is defined as the centroid of the ground truth mask. For clarity, we denote the Image Encoder as $IE$, the Box Predictor as $BP$, the Prompt Encoder as $PE$, and the Mask Decoder as $MD$ throughout this section.
	
	To simulate more practical interactive scenarios and potential inaccuracies during clicking, a random spatial perturbation $(\Delta x,\Delta y)$ is applied to the centroid. These displacements are uniformly sampled from the range $[-p,p]$, and the final sampled point is computed as:
    
     \begin{equation}
        \begin{aligned}
            p_{\text{sample}} &= \text{centroid} + (\Delta x, \Delta y), (\Delta x, \Delta y) &\sim \mathrm{Uniform}[-p, p]
        \end{aligned}
        \label{eq:perturbation}
    \end{equation}

\textcolor{black}{In Equation (\ref{eq:perturbation}), the perturbation limit 
$p$ is adapted to the scale of the anatomical structures within each dataset. 
Specifically, $p$ is set such that the resulting $2p \times 2p$ sampling 
region has an area equal to $30\%$ of the average mask area of the given 
dataset, i.e., $p = \frac{1}{2}\sqrt{0.3 \cdot \bar{A}}$, where $\bar{A}$ 
denotes the average mask area. The choice of $30\%$ is justified empirically 
through an ablation study presented in Section~\ref{sec:perturbation}, where 
it is shown to provide the optimal trade-off between robustness to off-center 
clicks and peak segmentation accuracy. This dynamic adjustment ensures that 
the magnitude of the simulated user error remains proportional to the object 
size, providing a realistic evaluation of robustness across different scales.} Although omitting the perturbation could increase accuracy under ideal conditions, it would reduce the model’s robustness in practical, interactive segmentation settings.

	When the perturbed point falls outside the mask boundary, a new point is randomly sampled from within the valid mask region to maintain data integrity. This approach guarantees that every training sample corresponds to a valid location within the target region, enhancing both stability and reliability during training.
	
	In the MedSAM architecture, each input image of size $1024\times1024$ produces an embedding tensor of shape $64\times64\times256$ through the Image Encoder. However, using the entire embedding space for bounding box prediction is both computationally inefficient and suboptimal, as it reduces the model’s focus on the localized area of interest. To address this, a localized patch surrounding the input point is extracted from the embedding tensor. Specifically, a $5\times5$ window centered around the prompt point is cropped and flattened into a feature vector. The $5\times5$ patch size is chosen as a trade-off: small enough to maintain spatial focus, yet large enough to capture local contextual cues from the surrounding tissue.
	
	\color{black}
	The flattened local embedding vector is then fed into a multi-layer perceptron (MLP) responsible for predicting the bounding box parameters. The MLP consists of three fully connected layers. The first layer with 256 neurons, the second layer with 128 neurons, and the output layer with 4 neurons, corresponding to the predicted width ($W$), height ($H$), and center offsets ($\Delta x, \Delta y$) relative to the input point.
	
	The three-layer design with 256 and 128 hidden units provides sufficient capacity to model the relationship between local embedding patterns and bounding box geometry while keeping the module lightweight; deeper architectures showed no performance gain in preliminary experiments.\color{black}
	
	The intermediate layers use ReLU activations, while a Softplus activation is applied to the width and height outputs to ensure positivity. This design enables the Box Predictor to efficiently produce approximate yet reliable bounding boxes using a lightweight architecture. Focusing on a local embedding patch not only reduces computational overhead but also enhances prediction accuracy.
	
	The MLP predicts four values: width, height, and relative center displacements ($\Delta x, \Delta y$). This allows the model to adapt both the size and position of the bounding box according to the underlying anatomical structure. For instance, when the input point is near the boundary of the target region, the model can shift the box center inward to better align with the actual object. This flexibility is essential for accommodating the varied shapes and sizes of anatomical structures in medical imaging.
	
	\subsection{Integration of the Box Predictor in the Training Pipeline}
	The integration of the Box Predictor into the second training stage aims to enhance the prompt representation and improve segmentation accuracy in MedSAM. At the beginning of this stage, the best-performing Box Predictor trained during the first stage is loaded and then kept frozen to prevent further updates. Likewise, the  Image Encoder is also kept frozen to preserve its pre-trained feature representations. This ensures that the learned visual knowledge and spatial understanding of both components remain stable. Only the Prompt Encoder and the Mask Decoder are fine-tuned during this stage.
	
	\subsubsection{Feature Extraction and Box Prediction (Frozen Modules)}
	Let $\Theta_{IE}$ and $\Theta_{BP}$ denote the weights of the Image Encoder and Box Predictor, respectively (as defined in Section 3.2). These parameters remain fixed throughout the process. The feature embedding is first extracted as:
	\begin{equation}
		F = IE(I; \Theta_{IE})
	\end{equation}
	where $I$ is the input image (resized to $1024\times1024$), and $F$ represents the resulting feature map of size $256\times64\times64$. Using the localized embedding patch $F_{localized}$ around the point prompt $P$, the Box Predictor estimates the bounding box $B$:
	\begin{equation}
		B = BP(P, F_{localized}; \Theta_{BP})
	\end{equation}
	The output $B$ consists of normalized coordinates $[x_{min}, y_{min}, x_{max}, y_{max}] \in [0,1]$. To ensure valid outputs, the raw predictions are processed through non-negative activations such as Softplus (for width and height) and clipped within the valid range to prevent numerical instability. Since MedSAM expects spatial prompts in pixel coordinates, the predicted normalized box $B$ is rescaled to the image resolution ($1024\times1024$) before being passed to the Prompt Encoder.
	
	\subsubsection{Prompt Encoding and Mask Generation (Trainable Modules)}
	The Prompt Encoder ($\Theta_{PE}$) and Mask Decoder ($\Theta_{MD}$) are the only trainable components in this stage. Let $P$ denote the point prompt provided by the user, and $B$ denote the bounding-box prompt predicted by the Box Predictor. The combined prompt $\{P,B\}$ is first embedded via:
	\begin{equation}
		E_P = PE(U; \Theta_{PE})
	\end{equation}
	Here, $U$ denotes the spatial prompt set, where $U=\{P,B\}$ when both point and box prompts are used, and $U=\{B\}$ when only the predicted box is provided. This operation produces a shared latent representation that captures the spatial extent defined by the box and, optionally, local positional information from the points. The Mask Decoder then combines this prompt embedding with the image features to generate the final predicted mask:
	\begin{equation}
		M = MD(E_P, F; \Theta_{MD})
	\end{equation}
	Using the box alone can already provide strong spatial guidance, while adding points can further refine localization in challenging regions. This flexible strategy allows the model to adapt to different input conditions and maintain stable and accurate segmentation, even under imperfect user interactions.
	
	\subsection{Loss Function}
	Our framework requires two distinct loss functions to train its components. The Box Predictor module is optimized using a bounding box regression loss, while the MedSAM-based segmentation model is trained using a mask prediction loss. These losses are applied independently to their respective modules during training.
	
	\subsubsection{Box Predictor Loss}
	The Box Predictor module is trained using a composite loss function that combines the L1 loss and the Generalized Intersection over Union (GIoU) \cite{12} loss. The overall objective function is defined as:
	\begin{equation}
		L = L_{L1} + L_{GIoU}
	\end{equation}
	The L1 loss measures the absolute difference between the predicted and ground truth bounding box coordinates:
	\begin{equation}
		L_{L1} = \frac{1}{N} \sum_{i=1}^{N} \|B_i - \hat{B}_i\|_1
	\end{equation}
	where $\hat{B}_i$ and $B_i$ denote the predicted and ground truth bounding boxes, respectively.
	
	In contrast, the GIoU loss is an extension of the classical Intersection over Union (IoU) metric, which additionally considers the smallest enclosing box that covers both the predicted and ground truth boxes. The GIoU is defined as:
	\begin{equation}
		GIoU = IoU - \frac{|C| - |B \cup \hat{B}|}{|C|}
	\end{equation}
	where $C$ denotes the smallest enclosing box, and $\hat{B}$ and $B$ correspond to the predicted and ground truth boxes, respectively. The GIoU loss is then computed as:
	\begin{equation}
		L_{GIoU} = 1 - GIoU
	\end{equation}
	The key advantage of GIoU over IoU lies in its ability to provide meaningful gradients even when there is no overlap between boxes ($IoU=0$). By incorporating the enclosing region, GIoU conveys information not only about overlap but also about the relative spatial alignment, distance, and direction of the predicted box with respect to the reference box. This geometric awareness helps the model learn precise spatial relationships rather than relying solely on the extent of overlap.
	
	Combining $L_{L1}$ and $L_{GIoU}$ yields an additive effect: the L1 term ensures coordinate-level precision in boundary localization, while the GIoU term promotes spatial consistency and geometric alignment. Such a dual objective is particularly beneficial for medical image segmentation, where both boundary accuracy and spatial coherence are critical for clinical analysis. All computations are performed on normalized coordinates within the [0,1] range to ensure numerical stability during optimization.
	
	\subsubsection{Segmentation Loss}
	For optimizing the MedSAM-based model, a hybrid loss function combining region-level and pixel-level objectives is adopted. The Dice loss promotes accurate overall overlap between predicted and ground truth masks, while the BCE loss enforces correctness at the individual pixel level. The total loss is defined as:
	\begin{equation}
		L_{total} = L_{DICE} + L_{BCE}
	\end{equation}
	The Dice loss measures the degree of overlap between the predicted mask and the ground truth mask, and is defined as:
	\begin{equation}
		L_{Dice} = 1 - \frac{2\sum_{i=1}^{N} p_i g_i + \epsilon}{\sum_{i=1}^{N} p_i^2 + \sum_{i=1}^{N} g_i^2 + \epsilon}
	\end{equation}
	where $p_i$ is the predicted probability for pixel $i$, $g_i$ is the corresponding ground truth label, and $\epsilon$ is a small constant to avoid division by zero.
	
	To complement this, the Binary Cross-Entropy (BCE) loss penalizes pixel-level misclassifications and is defined as:
	\begin{equation}
		L_{BCE} = -\frac{1}{N} \sum_{i=1}^{N} [g_i \log(p_i) + (1-g_i) \log(1-p_i)]
	\end{equation}
	Let $z_i$ denote the logit for pixel $i$, and $p_i = \sigma(z_i)$ its predicted probability where $\sigma$ denotes the sigmoid activation function. The combination of these two losses ensures both pixel-wise accuracy and region-wise consistency, leading to smoother and more anatomically stable segmentation results.

    \begin{figure*}[t]
		\centering
		
		\centering
		\includegraphics[width=0.8\linewidth]{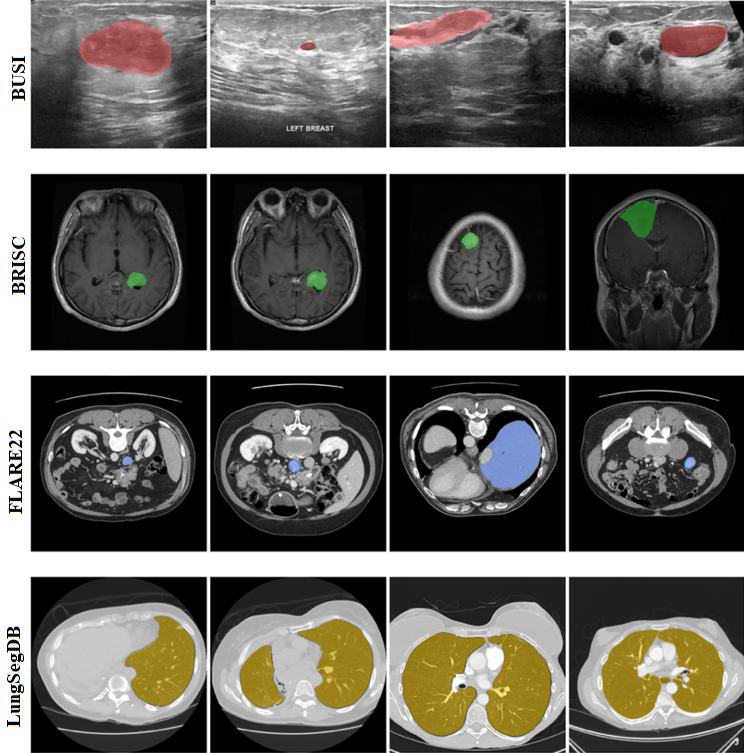}
		\caption{Representative examples from the four medical segmentation datasets. Each row shows four samples from one dataset with segmentation masks overlaid: BUSI (red, ultrasound), BRISC (green, brain MRI), FLARE22 (blue, abdominal CT), and LungSegDB (yellow, lung CT).}
		\label{fig:examples}
	\end{figure*}

	\section{Experimental Results}
	This section presents the experimental evaluation of the proposed method across four medical imaging datasets. To assess its effectiveness, the performance of our approach is compared against the baseline MedSAM under multiple experimental scenarios. Both quantitative and qualitative results are reported and analyzed to provide a comprehensive understanding of the model’s behavior in different imaging conditions.
	
	\subsection{Implementation Details}
	\label{sec:implementation_detail}

\textcolor{black}{All experiments were conducted on a single NVIDIA RTX 4090 GPU with PyTorch 2.9.1 (CUDA 12.8). The Box Predictor was trained for 10 epochs using the AdamW optimizer with a learning rate of $1\times10^{-4}$. In the second stage, the MedSAM Prompt Encoder and Mask Decoder were fine-tuned for 5 epochs using the same learning rate, while the Image Encoder and Box Predictor remained frozen. For all baselines and proposed variants (UNet, DeepLabV3+, SAM, SAM2.1, SAM2UNet, MedSAM2, MedSAM, and our method), images and masks were first resized to the target input resolution of the corresponding model before any subsequent preprocessing.} \textcolor{black}{All SAM-family methods were evaluated at $1024\times1024$, following the standard SAM/MedSAM input protocol. UNet and DeepLabV3+ were evaluated at both $1024\times1024$ and $512\times512$. The $1024\times1024$ setting enables a uniform input-resolution comparison with the SAM-family methods, whereas the $512\times512$ setting more closely matches the mean native image scale of BRISC ($\sim$505$\times$505), LungSegDB ($512\times512$), and BUSI ($581\times476$), providing a stronger reference for the CNN-based baselines. For these datasets, evaluation at $1024\times1024$ introduces an additional upsampling step that may alter the effective image scale and reduce CNN performance. In contrast, SAM- and MedSAM-based models require the $1024\times1024$ input protocol and therefore cannot be evaluated at the same lower resolution without changing their standard inference pipeline. Reporting UNet and DeepLabV3+ at both resolutions therefore provides both a uniform cross-method comparison and a more representative evaluation closer to the native image scale of these datasets.
}

\begin{table}[h]
	\centering
	\caption{\textcolor{black}{Summary of medical imaging datasets. The datasets span multiple modalities and anatomical regions to comprehensively evaluate segmentation performance.}}
	\label{tab:datasets}
	\renewcommand{\arraystretch}{1.3}
	\resizebox{\columnwidth}{!}{
		\begin{tabular}{cccccc}
			\hline
			\textcolor{black}{\textbf{Dataset}} &
			\textcolor{black}{\textbf{Modality}} &
			\textcolor{black}{\textbf{Region}} &
			\textcolor{black}{\textbf{Train}} &
			\textcolor{black}{\textbf{Test}} &
			\textcolor{black}{\textbf{Total}} \\
			\hline
			\textcolor{black}{FLARE22} &
			\textcolor{black}{CT} &
			\textcolor{black}{Abdomen} &
			\textcolor{black}{3000} &
			\textcolor{black}{621} &
			\textcolor{black}{3621} \\
			\textcolor{black}{BRISC} &
			\textcolor{black}{MRI} &
			\textcolor{black}{Brain} &
			\textcolor{black}{3933} &
			\textcolor{black}{860} &
			\textcolor{black}{4793} \\
			\textcolor{black}{BUSI} &
			\textcolor{black}{Ultrasound} &
			\textcolor{black}{Breast} &
			\textcolor{black}{678} &
			\textcolor{black}{251} &
			\textcolor{black}{929} \\
			\textcolor{black}{LungSegDB} &
			\textcolor{black}{CT} &
			\textcolor{black}{Lung} &
			\textcolor{black}{1371} &
			\textcolor{black}{343} &
			\textcolor{black}{1714} \\
			\hline
		\end{tabular}
	}
\end{table}

\begin{table}[t]
	\centering
	\caption{\textcolor{black}{Checkpoint and trainable-component summary for the evaluated baselines.}}
	\label{tab:protocol}
	\setlength{\tabcolsep}{4pt}
	\renewcommand{\arraystretch}{1.1}
	\resizebox{\columnwidth}{!}{%
		\begin{tabular}{lll}
			\toprule
			\textcolor{black}{\textbf{Method}} &
			\textcolor{black}{\textbf{Checkpoint / backbone}} &
			\textcolor{black}{\textbf{Trainable components}} \\
			\midrule
			\textcolor{black}{UNet} &
			\textcolor{black}{ResNet-50} &
			\textcolor{black}{All parameters} \\
			\textcolor{black}{DeepLabV3+} &
			\textcolor{black}{ResNet-50} &
			\textcolor{black}{All parameters} \\
			\textcolor{black}{SAM} &
			\textcolor{black}{ViT-B} &
			\textcolor{black}{Prompt encoder + mask decoder} \\
			\textcolor{black}{SAM2.1} &
			\textcolor{black}{Hiera-B+} &
			\textcolor{black}{Prompt encoder + mask decoder} \\
			\textcolor{black}{SAM2UNet} &
			\textcolor{black}{SAM2 Hiera-B+ encoder} &
			\textcolor{black}{Adapters + U-shaped decoder} \\
			\textcolor{black}{MedSAM2} &
			\textcolor{black}{SAM2.1-Hiera-Tiny} &
			\textcolor{black}{Prompt encoder + mask decoder} \\
			\textcolor{black}{MedSAM} &
			\textcolor{black}{ViT-B} &
			\textcolor{black}{Prompt encoder + mask decoder} \\
			\bottomrule
		\end{tabular}%
	}
\end{table}

	\subsection{Baseline Protocol}
\label{sec:baseline_protocol}

\textcolor{black}{All trainable baselines and proposed variants were trained or fine-tuned using the same training split reported in Table~\ref{tab:datasets}, and all results were computed on the corresponding test split. For all interactive prompt-based methods, the input point was generated using the same centroid-based perturbation policy described in Section~\ref{sec:method}, where the perturbation magnitude corresponds to $30\%$ of the average mask area. The same sampled point coordinates were reused across all point-prompted methods to ensure a fair comparison. All SAM-family methods were evaluated at $1024\times1024$, while UNet and DeepLabV3+ were evaluated at both $1024\times1024$ and $512\times512$. As reported in Table~\ref{tab:protocol}, UNet and DeepLabV3+ use ResNet-50 encoders initialized with ImageNet weights before training on our split. MedSAM2 uses the SAM2.1-Hiera-Tiny backbone, following the released MedSAM2 training configuration. This is analogous to MedSAM, which is released with a single ViT-B checkpoint. SAM2UNet is included as a prompt-free fully automatic SAM2-based baseline; therefore, it is not a one-to-one interactive comparison with the point-prompted methods.}

\begin{table*}[t]
\centering

\caption{\textcolor{black}{Segmentation performance across four datasets. Best results for each dataset and metric are shown in \textbf{bold}. R: Recall, P: Precision, IoU: Intersection over Union. UNet and DeepLabV3+ are evaluated at both $1024\times1024$ and $512\times512$, while all other methods are evaluated at $1024\times1024$. Interactive prompt-based methods (SAM, SAM2.1, MedSAM2, MedSAM + point, MedSAM + Box, and MedSAM + Box + Pt) are reported as mean $\pm$ standard deviation over three random seeds. Non-interactive methods (UNet, DeepLabV3+, and SAM2UNet) are reported as single runs.}}

\label{tab:comparison}
\setlength{\tabcolsep}{3pt}
\renewcommand{\arraystretch}{1.25}
\resizebox{\textwidth}{!}{%
\begin{tabular}{lcccccccccccccccc}
\toprule
\multirow{2}{*}{\textbf{Method}} &
\multicolumn{4}{c}{\textbf{BUSI}} &
\multicolumn{4}{c}{\textbf{FLARE22}} &
\multicolumn{4}{c}{\textbf{BRISC}} &
\multicolumn{4}{c}{\textbf{LungSegDB}} \\
\cmidrule(lr){2-5}\cmidrule(lr){6-9}\cmidrule(lr){10-13}\cmidrule(lr){14-17}
& R & P & IoU & Dice
& R & P & IoU & Dice
& R & P & IoU & Dice
& R & P & IoU & Dice \\
\midrule

UNet (1024$\times$1024)~\cite{19}
& \textcolor{black}{0.7536} & \textcolor{black}{0.5972} & \textcolor{black}{0.4709} & \textcolor{black}{0.5970}
& \textcolor{black}{0.3713} & \textcolor{black}{0.2211} & \textcolor{black}{0.1940} & \textcolor{black}{0.2569}
& \textcolor{black}{0.7604} & \textcolor{black}{0.8194} & \textcolor{black}{0.6730} & \textcolor{black}{0.7555}
& \textcolor{black}{0.9732} & \textcolor{black}{0.9813} & \textcolor{black}{0.9608} & \textcolor{black}{0.9765} \\[1pt]

\textcolor{black}{UNet (512$\times$512)~\cite{19}}
& \footnotesize\textcolor{black}{0.7414} & \footnotesize\textcolor{black}{0.7649} & \footnotesize\textcolor{black}{0.6207} & \footnotesize\textcolor{black}{0.7167}
& \footnotesize\textcolor{black}{0.3901} & \footnotesize\textcolor{black}{0.2293} & \footnotesize\textcolor{black}{0.2062} & \footnotesize\textcolor{black}{0.2706}
& \footnotesize\textcolor{black}{0.8294} & \footnotesize\textcolor{black}{0.8274} & \footnotesize\textcolor{black}{0.7233} & \footnotesize\textcolor{black}{0.8056}
& \footnotesize\textcolor{black}{0.9725} & \footnotesize\textcolor{black}{0.9781} & \footnotesize\textcolor{black}{0.9569} & \footnotesize\textcolor{black}{0.9746} \\[3pt]

DeepLabV3+ (1024$\times$1024)~\cite{18}
& \textcolor{black}{0.7548} & \textcolor{black}{0.7614} & \textcolor{black}{0.6195} & \textcolor{black}{0.7270}
& \textcolor{black}{0.3451} & \textcolor{black}{0.2166} & \textcolor{black}{0.1906} & \textcolor{black}{0.2515}
& \textcolor{black}{0.8622} & \textcolor{black}{0.8425} & \textcolor{black}{0.7553} & \textcolor{black}{0.8342}
& \textcolor{black}{0.9835} & \textcolor{black}{0.9807} & \textcolor{black}{0.9653} & \textcolor{black}{0.9812} \\[1pt]

\textcolor{black}{DeepLabV3+ (512$\times$512)~\cite{18}}
& \footnotesize\textcolor{black}{0.7677} & \footnotesize\textcolor{black}{0.8510} & \footnotesize\textcolor{black}{0.6868} & \footnotesize\textcolor{black}{0.7872}
& \footnotesize\textcolor{black}{0.4157} & \footnotesize\textcolor{black}{0.2280} & \footnotesize\textcolor{black}{0.1921} & \footnotesize\textcolor{black}{0.2683}
& \footnotesize\textcolor{black}{0.8443} & \footnotesize\textcolor{black}{0.8649} & \footnotesize\textcolor{black}{0.7588} & \footnotesize\textcolor{black}{0.8381}
& \footnotesize\textcolor{black}{0.9826} & \footnotesize\textcolor{black}{0.9812} & \footnotesize\textcolor{black}{0.9657} & \footnotesize\textcolor{black}{0.9811} \\[3pt]

\textcolor{black}{SAM2UNet}~\cite{38}
& \textcolor{black}{0.8975} & \textcolor{black}{0.8344} & \textcolor{black}{0.7703} & \textcolor{black}{0.8525}
& \textcolor{black}{0.4895} & \textcolor{black}{0.2355} & \textcolor{black}{0.1989} & \textcolor{black}{0.2777}
& \textbf{\textcolor{black}{0.9257}} & \textcolor{black}{0.8371} & \textcolor{black}{0.7907} & \textcolor{black}{0.8633}
& \textbf{\textcolor{black}{0.9890}} & \textcolor{black}{0.9850} & \textbf{\textcolor{black}{0.9748}} & \textbf{\textcolor{black}{0.9868}} \\[3pt]

\noalign{\vskip 4pt\hrule height 0.4pt\vskip 6pt}

SAM + point~\cite{2}
& \begin{tabular}[c]{@{}c@{}}\textcolor{black}{0.8580}\\\scriptsize{\textcolor{black}{($\pm$.0025)}}\end{tabular}
& \begin{tabular}[c]{@{}c@{}}\textcolor{black}{0.8326}\\\scriptsize{\textcolor{black}{($\pm$.0036)}}\end{tabular}
& \begin{tabular}[c]{@{}c@{}}\textcolor{black}{0.7219}\\\scriptsize{\textcolor{black}{($\pm$.0027)}}\end{tabular}
& \begin{tabular}[c]{@{}c@{}}\textcolor{black}{0.8251}\\\scriptsize{\textcolor{black}{($\pm$.0024)}}\end{tabular}
& \begin{tabular}[c]{@{}c@{}}\textcolor{black}{0.9179}\\\scriptsize{\textcolor{black}{($\pm$.0017)}}\end{tabular}
& \begin{tabular}[c]{@{}c@{}}\textcolor{black}{0.8367}\\\scriptsize{\textcolor{black}{($\pm$.0025)}}\end{tabular}
& \begin{tabular}[c]{@{}c@{}}\textcolor{black}{0.7825}\\\scriptsize{\textcolor{black}{($\pm$.0027)}}\end{tabular}
& \begin{tabular}[c]{@{}c@{}}\textcolor{black}{0.8503}\\\scriptsize{\textcolor{black}{($\pm$.0026)}}\end{tabular}
& \begin{tabular}[c]{@{}c@{}}\textcolor{black}{0.8741}\\\scriptsize{\textcolor{black}{($\pm$.0010)}}\end{tabular}
& \begin{tabular}[c]{@{}c@{}}\textcolor{black}{0.8718}\\\scriptsize{\textcolor{black}{($\pm$.0014)}}\end{tabular}
& \begin{tabular}[c]{@{}c@{}}\textcolor{black}{0.7679}\\\scriptsize{\textcolor{black}{($\pm$.0014)}}\end{tabular}
& \begin{tabular}[c]{@{}c@{}}\textcolor{black}{0.8502}\\\scriptsize{\textcolor{black}{($\pm$.0011)}}\end{tabular}
& \begin{tabular}[c]{@{}c@{}}\textcolor{black}{0.9711}\\\scriptsize{\textcolor{black}{($\pm$.0006)}}\end{tabular}
& \begin{tabular}[c]{@{}c@{}}\textcolor{black}{0.9808}\\\scriptsize{\textcolor{black}{($\pm$.0003)}}\end{tabular}
& \begin{tabular}[c]{@{}c@{}}\textcolor{black}{0.9578}\\\scriptsize{\textcolor{black}{($\pm$.0002)}}\end{tabular}
& \begin{tabular}[c]{@{}c@{}}\textcolor{black}{0.9751}\\\scriptsize{\textcolor{black}{($\pm$.0003)}}\end{tabular} \\[10pt]

\textcolor{black}{SAM2.1 + point}~\cite{40}
& \begin{tabular}[c]{@{}c@{}}\textcolor{black}{0.8657}\\\scriptsize{\textcolor{black}{($\pm$.0014)}}\end{tabular}
& \begin{tabular}[c]{@{}c@{}}\textbf{\textcolor{black}{0.9128}}\\\scriptsize{\textcolor{black}{($\pm$.0022)}}\end{tabular}
& \begin{tabular}[c]{@{}c@{}}\textcolor{black}{0.7957}\\\scriptsize{\textcolor{black}{($\pm$.0023)}}\end{tabular}
& \begin{tabular}[c]{@{}c@{}}\textcolor{black}{0.8789}\\\scriptsize{\textcolor{black}{($\pm$.0018)}}\end{tabular}
& \begin{tabular}[c]{@{}c@{}}\textcolor{black}{0.9011}\\\scriptsize{\textcolor{black}{($\pm$.0020)}}\end{tabular}
& \begin{tabular}[c]{@{}c@{}}\textcolor{black}{0.8918}\\\scriptsize{\textcolor{black}{($\pm$.0013)}}\end{tabular}
& \begin{tabular}[c]{@{}c@{}}\textcolor{black}{0.8209}\\\scriptsize{\textcolor{black}{($\pm$.0011)}}\end{tabular}
& \begin{tabular}[c]{@{}c@{}}\textcolor{black}{0.8806}\\\scriptsize{\textcolor{black}{($\pm$.0015)}}\end{tabular}
& \begin{tabular}[c]{@{}c@{}}\textcolor{black}{0.9038}\\\scriptsize{\textcolor{black}{($\pm$.0010)}}\end{tabular}
& \begin{tabular}[c]{@{}c@{}}\textcolor{black}{0.8710}\\\scriptsize{\textcolor{black}{($\pm$.0016)}}\end{tabular}
& \begin{tabular}[c]{@{}c@{}}\textcolor{black}{0.7936}\\\scriptsize{\textcolor{black}{($\pm$.0016)}}\end{tabular}
& \begin{tabular}[c]{@{}c@{}}\textcolor{black}{0.8695}\\\scriptsize{\textcolor{black}{($\pm$.0015)}}\end{tabular}
& \begin{tabular}[c]{@{}c@{}}\textcolor{black}{0.9813}\\\scriptsize{\textcolor{black}{($\pm$.0002)}}\end{tabular}
& \begin{tabular}[c]{@{}c@{}}\textcolor{black}{0.9811}\\\scriptsize{\textcolor{black}{($\pm$.0001)}}\end{tabular}
& \begin{tabular}[c]{@{}c@{}}\textcolor{black}{0.9659}\\\scriptsize{\textcolor{black}{($\pm$.0003)}}\end{tabular}
& \begin{tabular}[c]{@{}c@{}}\textcolor{black}{0.9806}\\\scriptsize{\textcolor{black}{($\pm$.0002)}}\end{tabular} \\[10pt]

\textcolor{black}{MedSAM2 + point}~\cite{39}
& \begin{tabular}[c]{@{}c@{}}\textcolor{black}{0.8503}\\\scriptsize{\textcolor{black}{($\pm$.0007)}}\end{tabular}
& \begin{tabular}[c]{@{}c@{}}\textcolor{black}{0.8534}\\\scriptsize{\textcolor{black}{($\pm$.0008)}}\end{tabular}
& \begin{tabular}[c]{@{}c@{}}\textcolor{black}{0.7304}\\\scriptsize{\textcolor{black}{($\pm$.0022)}}\end{tabular}
& \begin{tabular}[c]{@{}c@{}}\textcolor{black}{0.8348}\\\scriptsize{\textcolor{black}{($\pm$.0015)}}\end{tabular}
& \begin{tabular}[c]{@{}c@{}}\textcolor{black}{0.9058}\\\scriptsize{\textcolor{black}{($\pm$.0018)}}\end{tabular}
& \begin{tabular}[c]{@{}c@{}}\textcolor{black}{0.9049}\\\scriptsize{\textcolor{black}{($\pm$.0015)}}\end{tabular}
& \begin{tabular}[c]{@{}c@{}}\textcolor{black}{0.8379}\\\scriptsize{\textcolor{black}{($\pm$.0015)}}\end{tabular}
& \begin{tabular}[c]{@{}c@{}}\textcolor{black}{0.8964}\\\scriptsize{\textcolor{black}{($\pm$.0013)}}\end{tabular}
& \begin{tabular}[c]{@{}c@{}}\textcolor{black}{0.8627}\\\scriptsize{\textcolor{black}{($\pm$.0007)}}\end{tabular}
& \begin{tabular}[c]{@{}c@{}}\textcolor{black}{0.8872}\\\scriptsize{\textcolor{black}{($\pm$.0011)}}\end{tabular}
& \begin{tabular}[c]{@{}c@{}}\textcolor{black}{0.7729}\\\scriptsize{\textcolor{black}{($\pm$.0002)}}\end{tabular}
& \begin{tabular}[c]{@{}c@{}}\textcolor{black}{0.8562}\\\scriptsize{\textcolor{black}{($\pm$.0003)}}\end{tabular}
& \begin{tabular}[c]{@{}c@{}}\textcolor{black}{0.9850}\\\scriptsize{\textcolor{black}{($\pm$.0007)}}\end{tabular}
& \begin{tabular}[c]{@{}c@{}}\textcolor{black}{0.9812}\\\scriptsize{\textcolor{black}{($\pm$.0003)}}\end{tabular}
& \begin{tabular}[c]{@{}c@{}}\textcolor{black}{0.9689}\\\scriptsize{\textcolor{black}{($\pm$.0002)}}\end{tabular}
& \begin{tabular}[c]{@{}c@{}}\textcolor{black}{0.9828}\\\scriptsize{\textcolor{black}{($\pm$.0005)}}\end{tabular} \\

\noalign{\vskip 4pt\hrule height 0.4pt\vskip 6pt}

MedSAM + point~\cite{3}
& \begin{tabular}[c]{@{}c@{}}0.8945\\\scriptsize{\textcolor{black}{($\pm$.0070)}}\end{tabular}
& \begin{tabular}[c]{@{}c@{}}0.8843\\\scriptsize{\textcolor{black}{($\pm$.0010)}}\end{tabular}
& \begin{tabular}[c]{@{}c@{}}0.7936\\\scriptsize{\textcolor{black}{($\pm$.0054)}}\end{tabular}
& \begin{tabular}[c]{@{}c@{}}0.8771\\\scriptsize{\textcolor{black}{($\pm$.0042)}}\end{tabular}
& \begin{tabular}[c]{@{}c@{}}0.9262\\\scriptsize{\textcolor{black}{($\pm$.0019)}}\end{tabular}
& \begin{tabular}[c]{@{}c@{}}0.9214\\\scriptsize{\textcolor{black}{($\pm$.0018)}}\end{tabular}
& \begin{tabular}[c]{@{}c@{}}0.8692\\\scriptsize{\textcolor{black}{($\pm$.0014)}}\end{tabular}
& \begin{tabular}[c]{@{}c@{}}0.9169\\\scriptsize{\textcolor{black}{($\pm$.0018)}}\end{tabular}
& \begin{tabular}[c]{@{}c@{}}0.8860\\\scriptsize{\textcolor{black}{($\pm$.0015)}}\end{tabular}
& \begin{tabular}[c]{@{}c@{}}\textbf{0.8959}\\\scriptsize{\textcolor{black}{($\pm$.0009)}}\end{tabular}
& \begin{tabular}[c]{@{}c@{}}0.7978\\\scriptsize{\textcolor{black}{($\pm$.0008)}}\end{tabular}
& \begin{tabular}[c]{@{}c@{}}0.8745\\\scriptsize{\textcolor{black}{($\pm$.0008)}}\end{tabular}
& \begin{tabular}[c]{@{}c@{}}0.9786\\\scriptsize{\textcolor{black}{($\pm$.0008)}}\end{tabular}
& \begin{tabular}[c]{@{}c@{}}0.9816\\\scriptsize{\textcolor{black}{($\pm$.0002)}}\end{tabular}
& \begin{tabular}[c]{@{}c@{}}0.9626\\\scriptsize{\textcolor{black}{($\pm$.0006)}}\end{tabular}
& \begin{tabular}[c]{@{}c@{}}0.9788\\\scriptsize{\textcolor{black}{($\pm$.0005)}}\end{tabular} \\[10pt]

MedSAM + Box \textit{(ours)}
& \begin{tabular}[c]{@{}c@{}}\textbf{0.9176}\\\scriptsize{\textcolor{black}{($\pm$.0028)}}\end{tabular}
& \begin{tabular}[c]{@{}c@{}}0.8814\\\scriptsize{\textcolor{black}{($\pm$.0003)}}\end{tabular}
& \begin{tabular}[c]{@{}c@{}}\textbf{0.8124}\\\scriptsize{\textcolor{black}{($\pm$.0031)}}\end{tabular}
& \begin{tabular}[c]{@{}c@{}}\textbf{0.8904}\\\scriptsize{\textcolor{black}{($\pm$.0026)}}\end{tabular}
& \begin{tabular}[c]{@{}c@{}}\textbf{0.9492}\\\scriptsize{\textcolor{black}{($\pm$.0014)}}\end{tabular}
& \begin{tabular}[c]{@{}c@{}}0.9256\\\scriptsize{\textcolor{black}{($\pm$.0011)}}\end{tabular}
& \begin{tabular}[c]{@{}c@{}}\textbf{0.8874}\\\scriptsize{\textcolor{black}{($\pm$.0014)}}\end{tabular}
& \begin{tabular}[c]{@{}c@{}}\textbf{0.9314}\\\scriptsize{\textcolor{black}{($\pm$.0012)}}\end{tabular}
& \begin{tabular}[c]{@{}c@{}}0.9106\\\scriptsize{\textcolor{black}{($\pm$.0008)}}\end{tabular}
& \begin{tabular}[c]{@{}c@{}}0.8770\\\scriptsize{\textcolor{black}{($\pm$.0017)}}\end{tabular}
& \begin{tabular}[c]{@{}c@{}}0.8035\\\scriptsize{\textcolor{black}{($\pm$.0008)}}\end{tabular}
& \begin{tabular}[c]{@{}c@{}}0.8793\\\scriptsize{\textcolor{black}{($\pm$.0007)}}\end{tabular}
& \begin{tabular}[c]{@{}c@{}}0.9812\\\scriptsize{\textcolor{black}{($\pm$.0002)}}\end{tabular}
& \begin{tabular}[c]{@{}c@{}}0.9839\\\scriptsize{\textcolor{black}{($\pm$.0001)}}\end{tabular}
& \begin{tabular}[c]{@{}c@{}}0.9664\\\scriptsize{\textcolor{black}{($\pm$.0001)}}\end{tabular}
& \begin{tabular}[c]{@{}c@{}}0.9816\\\scriptsize{\textcolor{black}{($\pm$.0000)}}\end{tabular} \\[10pt]

MedSAM + Box + Pt \textit{(ours)}
& \begin{tabular}[c]{@{}c@{}}0.9043\\\scriptsize{\textcolor{black}{($\pm$.0047)}}\end{tabular}
& \begin{tabular}[c]{@{}c@{}}0.8911\\\scriptsize{\textcolor{black}{($\pm$.0005)}}\end{tabular}
& \begin{tabular}[c]{@{}c@{}}0.8105\\\scriptsize{\textcolor{black}{($\pm$.0042)}}\end{tabular}
& \begin{tabular}[c]{@{}c@{}}0.8893\\\scriptsize{\textcolor{black}{($\pm$.0039)}}\end{tabular}
& \begin{tabular}[c]{@{}c@{}}0.9353\\\scriptsize{\textcolor{black}{($\pm$.0029)}}\end{tabular}
& \begin{tabular}[c]{@{}c@{}}\textbf{0.9366}\\\scriptsize{\textcolor{black}{($\pm$.0020)}}\end{tabular}
& \begin{tabular}[c]{@{}c@{}}0.8868\\\scriptsize{\textcolor{black}{($\pm$.0028)}}\end{tabular}
& \begin{tabular}[c]{@{}c@{}}0.9307\\\scriptsize{\textcolor{black}{($\pm$.0025)}}\end{tabular}
& \begin{tabular}[c]{@{}c@{}}0.8992\\\scriptsize{\textcolor{black}{($\pm$.0014)}}\end{tabular}
& \begin{tabular}[c]{@{}c@{}}0.8903\\\scriptsize{\textcolor{black}{($\pm$.0029)}}\end{tabular}
& \begin{tabular}[c]{@{}c@{}}\textbf{0.8059}\\\scriptsize{\textcolor{black}{($\pm$.0012)}}\end{tabular}
& \begin{tabular}[c]{@{}c@{}}\textbf{0.8815}\\\scriptsize{\textcolor{black}{($\pm$.0011)}}\end{tabular}
& \begin{tabular}[c]{@{}c@{}}0.9785\\\scriptsize{\textcolor{black}{($\pm$.0002)}}\end{tabular}
& \begin{tabular}[c]{@{}c@{}}\textbf{0.9868}\\\scriptsize{\textcolor{black}{($\pm$.0002)}}\end{tabular}
& \begin{tabular}[c]{@{}c@{}}0.9663\\\scriptsize{\textcolor{black}{($\pm$.0004)}}\end{tabular}
& \begin{tabular}[c]{@{}c@{}}0.9816\\\scriptsize{\textcolor{black}{($\pm$.0002)}}\end{tabular} \\

\bottomrule
\end{tabular}%
}
\end{table*}

\begin{table*}[t]
	\centering
	\caption{\textcolor{black}{Paired bootstrap analysis ($B=10{,}000$) over per-image Dice scores. Confidence intervals (95\%) are computed for the mean paired Dice difference between the proposed method and representative baselines. Confidence intervals excluding zero indicate statistically significant differences.}}
	\label{tab:bootstrap}
	\setlength{\tabcolsep}{6pt}
	\renewcommand{\arraystretch}{1.2}
	\resizebox{\textwidth}{!}{%
		\begin{tabular}{lllcc}
			\toprule
			\textcolor{black}{\textbf{Dataset}} &
			\textcolor{black}{\textbf{Proposed Method}} &
			\textcolor{black}{\textbf{Compared Method}} &
			\textcolor{black}{\textbf{95\% Bootstrap CI}} &
			\textcolor{black}{\textbf{Interpretation}} \\
			\midrule
			
			\multirow{5}{*}{\textcolor{black}{BUSI}}
			& \multirow{5}{*}{\textcolor{black}{MedSAM + Box}}
			& \textcolor{black}{MedSAM + point} & \textcolor{black}{[+0.0045, +0.0226]} & \textcolor{black}{Significant} \\
			&
			& \textcolor{black}{MedSAM2} & \textcolor{black}{[+0.0441, +0.0672]} & \textcolor{black}{Significant} \\
			&
			& \textcolor{black}{SAM} & \textcolor{black}{[+0.0517, +0.0793]} & \textcolor{black}{Significant} \\
			&
			& \textcolor{black}{SAM2} & \textcolor{black}{[+0.0012, +0.0221]} & \textcolor{black}{Significant} \\
			&
			& \textcolor{black}{SAM2UNet} & \textcolor{black}{[+0.0194, +0.0575]} & \textcolor{black}{Significant} \\
			
			\midrule
			
			\multirow{5}{*}{\textcolor{black}{FLARE22}}
			& \multirow{5}{*}{\textcolor{black}{MedSAM + Box}}
			& \textcolor{black}{MedSAM + point} & \textcolor{black}{[+0.0048, +0.0246]} & \textcolor{black}{Significant} \\
			&
			& \textcolor{black}{MedSAM2} & \textcolor{black}{[+0.0259, +0.0444]} & \textcolor{black}{Significant} \\
			&
			& \textcolor{black}{SAM} & \textcolor{black}{[+0.0700, +0.0924]} & \textcolor{black}{Significant} \\
			&
			& \textcolor{black}{SAM2} & \textcolor{black}{[+0.0409, +0.0610]} & \textcolor{black}{Significant} \\
			&
			& \textcolor{black}{SAM2UNet} & \textcolor{black}{[+0.6312, +0.6753]} & \textcolor{black}{Significant} \\
			
			\midrule
			
			\multirow{5}{*}{\textcolor{black}{BRISC}}
			& \multirow{5}{*}{\textcolor{black}{MedSAM + Box}}
			& \textcolor{black}{MedSAM + point} & \textcolor{black}{[+0.0019, +0.0121]} & \textcolor{black}{Significant} \\
			&
			& \textcolor{black}{MedSAM2} & \textcolor{black}{[+0.0186, +0.0318]} & \textcolor{black}{Significant} \\
			&
			& \textcolor{black}{SAM} & \textcolor{black}{[+0.0244, +0.0386]} & \textcolor{black}{Significant} \\
			&
			& \textcolor{black}{SAM2} & \textcolor{black}{[+0.0054, +0.0185]} & \textcolor{black}{Significant} \\
			&
			& \textcolor{black}{SAM2UNet} & \textcolor{black}{[+0.0094, +0.0271]} & \textcolor{black}{Significant} \\
			
			\midrule
			
			\multirow{5}{*}{\textcolor{black}{LungSegDB}}
			& \multirow{5}{*}{\textcolor{black}{MedSAM + Box}}
			& \textcolor{black}{MedSAM + point} & \textcolor{black}{[+0.0007, +0.0061]} & \textcolor{black}{Significant (small effect)} \\
			&
			& \textcolor{black}{MedSAM2} & \textcolor{black}{[-0.0021, +0.0000]} & \textcolor{black}{Not significant} \\
			&
			& \textcolor{black}{SAM} & \textcolor{black}{[+0.0031, +0.0106]} & \textcolor{black}{Significant (small effect)} \\
			&
			& \textcolor{black}{SAM2} & \textcolor{black}{[-0.0007, +0.0030]} & \textcolor{black}{Not significant} \\
			&
			& \textcolor{black}{SAM2UNet} & \textcolor{black}{[-0.0081, -0.0032]} & \textcolor{black}{SAM2UNet significantly better} \\
			
			\bottomrule
		\end{tabular}%
	}
\end{table*}

	\subsection{Datasets}
	To evaluate the proposed method, four publicly available medical imaging datasets were used. These datasets were chosen to cover a broad range of anatomical structures, imaging modalities, and segmentation challenges, ensuring that the experiments reflect diverse real-world clinical scenarios.

	\subsubsection{FLARE22}
	The FLARE22 dataset comprises abdominal CT scans from 2,900 patients across 53 multinational sites, covering 13 organs, multiple cancers, phases, and scanners. Since our method operates on 2D slices, all slices containing at least one annotated structure were extracted to form a 2D dataset, resulting in 3,621 usable samples. The dataset contains several large abdominal organs with substantial variation in size and position, making it particularly challenging for prompt-based segmentation methods \cite{14}.
	
	\subsubsection{BUSI}
	The BUSI (Breast Ultrasound Images) dataset includes ultrasound images collected from 600 female subjects aged 25–75. The dataset covers benign tumors, malignant tumors, and normal tissue. Due to the noisy nature of ultrasound and the presence of irregular lesion boundaries, BUSI provides a challenging testbed for segmentation models, especially in settings involving automatic prompt generation \cite{15}.
	
	\subsubsection{BRISC}
	The BRISC dataset contains contrast-enhanced T1-weighted brain MRI images annotated for four categories: glioma, meningioma, pituitary tumor, and no-tumor. All annotations were manually created by expert radiologists. The dataset encompasses a wide variety of tumor shapes, sizes, and anatomical locations and includes images in axial, sagittal, and coronal planes. This variability makes BRISC suitable for assessing the robustness of segmentation models under complex structural variations \cite{16}.
	
	\subsubsection{LungSegDB}
	LungSegDB is a curated collection of CT images of the lungs, compiled from multiple trusted sources. The dataset provides manually generated segmentation masks aimed at supporting automated detection of pulmonary conditions such as cancer and COVID-19. The relatively clean imaging modality combined with well-defined organ boundaries makes LungSegDB a suitable dataset for evaluating segmentation performance when the underlying structures are less ambiguous \cite{17}.
	
	\subsection{Evaluation Metrics}

To quantitatively evaluate segmentation performance, we employ standard region-based metrics commonly used in medical image analysis, including the Dice Similarity Coefficient (Dice), Intersection over Union (IoU), Precision, and Recall.  \textcolor{black}{These metrics measure the spatial agreement between the predicted segmentation mask $\hat{M}$ and the ground truth annotation $M$.}

The Dice score is defined as:

\begin{equation}
\textcolor{black}{\text{Dice} = \frac{2 \lvert \hat{M} \cap M \rvert}{\lvert \hat{M} \rvert + \lvert M \rvert}}
\end{equation}

The Intersection over Union (IoU) is given by:

\begin{equation}
\textcolor{black}{\text{IoU} = \frac{\lvert \hat{M} \cap M \rvert}{\lvert \hat{M} \cup M \rvert}}
\end{equation}

Precision and Recall evaluate the correctness and completeness of the predicted segmentation, respectively, and are defined as:

\begin{equation}
\textcolor{black}{\text{Precision} = \frac{\lvert \hat{M} \cap M \rvert}{\lvert \hat{M} \rvert}}
\end{equation}

\begin{equation}
\textcolor{black}{\text{Recall} = \frac{\lvert \hat{M} \cap M \rvert}{\lvert M \rvert}}
\end{equation}
All metrics take values in the range $[0,1]$, where higher values indicate better segmentation performance. Dice and IoU assess overall spatial overlap, while Precision and Recall characterize false-positive and false-negative behavior, respectively. All metrics are computed per image and averaged across the test set to obtain the final reported values. Together, these metrics provide a comprehensive evaluation of segmentation quality across different experimental settings.

\subsection{Comparison with Baseline Approaches}
\label{sec:comparison}

\textcolor{black}{To evaluate the proposed method, experiments were conducted on four medical imaging datasets: BRISC, FLARE22, LungSegDB, and BUSI. Performance was measured using Dice, IoU, Precision, and Recall. All interactive baselines are evaluated under a fixed, model-independent prompting protocol, where a single foreground point per image is sampled from the mask centroid with perturbation magnitude corresponding to $30\%$ of the average mask area, and the same coordinates are reused across all prompt-based methods. Quantitative results are summarized in Table~\ref{tab:comparison}.}

\textcolor{black}{To support head-to-head comparisons beyond seed-level variance, we conducted paired bootstrap tests ($B=10{,}000$) over per-image Dice scores on each test set. For each comparison, images were resampled with replacement while preserving the pairing between methods, and the 95\% confidence interval was computed for the mean paired Dice difference. The complete statistical comparison results are summarized in Table~\ref{tab:bootstrap}.}

\textcolor{black}{On BUSI, paired bootstrap analysis (Table~\ref{tab:bootstrap}) shows that MedSAM + Box significantly outperforms all compared baselines. These results suggest that the predicted box prompt helps capture lesion extent and reduce false negatives, which are common in ultrasound images with weak or irregular boundaries.}

\textcolor{black}{On FLARE22, MedSAM + Box also achieves statistically significant improvements over all compared methods (Table~\ref{tab:bootstrap}). The particularly large improvement over the prompt-free SAM2UNet baseline highlights the difficulty of multi-organ CT segmentation without explicit spatial guidance. These findings indicate that predicted box prompts provide effective spatial priors for complex anatomical structures.}

\textcolor{black}{On BRISC, paired bootstrap analysis confirms that MedSAM + Box + Pt significantly outperforms all compared baselines (Table~\ref{tab:bootstrap}). This indicates that combining box and point prompts provides complementary information, leading to more accurate brain tumor segmentation where lesion boundaries are often ambiguous.}

\textcolor{black}{On LungSegDB, all methods approach ceiling performance, reflecting the well-defined lung anatomy in CT imaging. As summarized in Table~\ref{tab:bootstrap}, MedSAM + Box is not statistically distinguishable from MedSAM2 or SAM2, while its improvements over MedSAM + point and SAM are statistically significant but very small, indicating only a marginal advantage in this near-saturated setting. In contrast, SAM2UNet achieves a significantly higher Dice score than MedSAM + Box, suggesting that its prompt-free design is particularly well suited to this anatomically well-defined segmentation task.}

\textcolor{black}{A comparison of the CNN baselines across the two evaluated input resolutions shows that the $512\times512$ setting yields better performance on datasets with smaller native image sizes, consistent with the additional upsampling step introduced at $1024\times1024$. Nevertheless, even at the $512\times512$ setting, the CNN baselines remain below the MedSAM-based promptable methods on the more challenging datasets, particularly BUSI and FLARE22. This indicates that the performance gap reflects the benefits of foundation-model pretraining and prompt-based spatial guidance rather than the upsampling-related resolution mismatch alone.}

\textcolor{black}{Among the additional baselines, SAM2.1 achieves competitive results across datasets, with performance close to MedSAM-based methods despite not being fine-tuned on medical data. SAM2UNet, as a non-interactive fully automatic segmentation method, does not rely on spatial prompts at inference time, which allows it to perform well on datasets where the target structure is visually unambiguous. However, on FLARE22, which contains multiple abdominal organs with high variability in size, position, and appearance across CT slices, SAM2UNet achieves a Dice score of only 0.2777. This reduced performance illustrates a limitation of prompt-free approaches in complex multi-organ segmentation: without spatial guidance to specify the intended structure, the model cannot reliably localize the correct target among many possible candidates.}

\textcolor{black}{SAM2-based models do not consistently outperform their SAM1-based counterparts in our experiments. This can be understood from an architectural perspective: SAM2's design improvements primarily target temporal consistency across video frames and volumetric sequences, whereas our evaluation is performed on independent 2D slices. Therefore, these temporal advantages are not fully exploited. This explains why MedSAM, despite being based on SAM1, achieves stronger results than MedSAM2 across most datasets: the domain fine-tuning in MedSAM is directly effective in the 2D setting, whereas MedSAM2's temporal design benefits remain largely inactive. \cite{41} reports that SAM2 offers no consistent advantage over SAM1 on 2D medical inputs, and the benchmark in \cite{43} evaluating multiple SAM variants across imaging modalities similarly shows that MedSAM achieves higher scores than SAM2-based variants.}

\subsection{Ablation Studies}
	To understand the contribution of individual components and design choices in our framework, we conducted comprehensive ablation experiments.
	
\begin{table*}[t]
    \centering
    \caption{Ablation study on MedSAM configurations using Dice (\%). The Image 
    Encoder is frozen in all experiments. `Trainable' indicates whether the SAM 
    Prompt Encoder and Mask Decoder are fine-tuned. \textcolor{black}{Values reported as mean\,$\pm$\,std over three seeds.}}
    \label{tab:ablation}
    
    \renewcommand{\arraystretch}{1.2}
    
    \begin{tabular}{lcc cccc c cc}
        \toprule
        
        \multicolumn{3}{c}{\textbf{Configuration}} & 
        \multicolumn{4}{c}{\textbf{Dataset Performance (Dice)}} & 
        \multirow{2}{*}{{\shortstack{\textbf{\textcolor{black}{Inference}}\\\textbf{\textcolor{black}{Time (ms)}}}}} & 
        \multicolumn{2}{c}{\textbf{\textcolor{black}{Trainable Params (M)}}} \\
        \cmidrule(r){1-3} \cmidrule(lr){4-7} \cmidrule(lr){9-10}
        
        \textbf{Decoder Mode} & \textbf{Point} & \textbf{Box} & 
        \textbf{BRISC} & \textbf{LungSegDB} & \textbf{BUSI} & \textbf{FLARE22} & &
        \textbf{\textcolor{black}{Decoder}} & \textbf{\textcolor{black}{Box Pred.}} \\
        \midrule
        \multirow{4}{*}{Frozen} 
          & \checkmark & $\times$
          & $0.1053{\scriptstyle\,\pm\textcolor{black}{0.0022}}$
          & $0.0164{\scriptstyle\,\pm\textcolor{black}{0.0007}}$
          & $0.0804{\scriptstyle\,\pm\textcolor{black}{0.0039}}$
          & $0.0568{\scriptstyle\,\pm\textcolor{black}{0.0043}}$
          & \textcolor{black}{61.49} & \textcolor{black}{0} & \textcolor{black}{0} \\
          & $\times$ & \checkmark
          & $0.8378{\scriptstyle\,\pm\textcolor{black}{0.0010}}$
          & $0.7258{\scriptstyle\,\pm\textcolor{black}{0.0023}}$
          & $0.7632{\scriptstyle\,\pm\textcolor{black}{0.0054}}$
          & $0.8905{\scriptstyle\,\pm\textcolor{black}{0.0015}}$
          & \textcolor{black}{62.30} & \textcolor{black}{0} & \textcolor{black}{1.6} \\
          & \checkmark & \checkmark
          & $0.7341{\scriptstyle\,\pm\textcolor{black}{0.0010}}$
          & $0.5006{\scriptstyle\,\pm\textcolor{black}{0.0077}}$
          & $0.5743{\scriptstyle\,\pm\textcolor{black}{0.0216}}$
          & $0.8284{\scriptstyle\,\pm\textcolor{black}{0.0021}}$
          & \textcolor{black}{62.57} & \textcolor{black}{0} & \textcolor{black}{1.6} \\
          & \multicolumn{2}{c}{{\footnotesize \textcolor{black}{Perfect Box}}}
          & \textcolor{black}{0.8950} & \textcolor{black}{0.9566} & \textcolor{black}{0.9259} & \textcolor{black}{0.9392}
          & \textcolor{black}{--} & \textcolor{black}{0} & \textcolor{black}{0} \\
        \midrule
        \multirow{4}{*}{Trainable}
          & \checkmark & $\times$
          & $0.8745{\scriptstyle\,\pm\textcolor{black}{0.0008}}$
          & $0.9788{\scriptstyle\,\pm\textcolor{black}{0.0005}}$
          & $0.8771{\scriptstyle\,\pm\textcolor{black}{0.0042}}$
          & $0.9169{\scriptstyle\,\pm\textcolor{black}{0.0018}}$
          & \textcolor{black}{61.53} & \textcolor{black}{4} & \textcolor{black}{0} \\
          & $\times$ & \checkmark
          & $0.8793{\scriptstyle\,\pm\textcolor{black}{0.0007}}$
          & $\mathbf{0.9816}{\scriptstyle\,\pm\textcolor{black}{0.0000}}$
          & $\mathbf{0.8904}{\scriptstyle\,\pm\textcolor{black}{0.0026}}$
          & $\mathbf{0.9314}{\scriptstyle\,\pm\textcolor{black}{0.0012}}$
          & \textcolor{black}{62.24} & \textcolor{black}{4} & \textcolor{black}{1.6} \\
          & \checkmark & \checkmark
          & $\mathbf{0.8815}{\scriptstyle\,\pm\textcolor{black}{0.0011}}$
          & $\mathbf{0.9816}{\scriptstyle\,\pm\textcolor{black}{0.0002}}$
          & $0.8893{\scriptstyle\,\pm\textcolor{black}{0.0039}}$
          & $0.9307{\scriptstyle\,\pm\textcolor{black}{0.0025}}$
          & \textcolor{black}{62.58} & \textcolor{black}{4} & \textcolor{black}{1.6} \\
          & \multicolumn{2}{c}{{\footnotesize \textcolor{black}{Perfect Box}}}
          & \textbf{\textcolor{black}{0.9320}} & \textbf{\textcolor{black}{0.9855}} & \textbf{\textcolor{black}{0.9576}} & \textbf{\textcolor{black}{0.9603}}
          & \textcolor{black}{--} & \textcolor{black}{4} & \textcolor{black}{0} \\
        \bottomrule
    \end{tabular}
\end{table*}

    \subsubsection{Impact of the Box Predictor Module}
    To evaluate the contribution of the Box Predictor, we tested six configurations of MedSAM. In the frozen settings, MedSAM was used with its pretrained weights and received either a point prompt, a predicted bounding box, or both, allowing us to assess how well the Box Predictor can assist the model without any parameter updates. In the trainable settings, the prompt encoder and mask decoder were fine-tuned using either point prompts, predicted boxes, or their combination, enabling analysis of how the module interacts with the model when adaptation to the target data is allowed. Together, these configurations provide a structured way to isolate the effect of the Box Predictor under both fixed-weight and fine-tuned conditions while covering all relevant prompt combinations.

    As shown in Table \ref{tab:ablation}, when MedSAM is used without any retraining, relying solely on a point prompt results in very poor performance across all datasets (Dice scores ranging from 1.64 to 10.53). Introducing the Box Predictor substantially changes this behavior. \textcolor{black}{While the Box Predictor achieves solid regression performance across all datasets (as reported in Table~\ref{tab:box_metrics}), the significant performance recovery is not solely a consequence of box quality. More fundamentally, it reflects the restoration of the geometric prior that MedSAM was conditioned on during fine-tuning. Since MedSAM was fine-tuned with bounding box prompts as the primary prompt type, its prompt encoder and mask decoder have learned to expect spatial inputs in this form. Providing a predicted bounding box, even one that is not perfectly tight, re-establishes this geometric conditioning, allowing the frozen model to recover its intended segmentation behavior.} \textcolor{black}{Beyond the fine-tuning mismatch, the severity of this collapse is also explained by the inherent ambiguity of point prompts. As reported in \cite{42}, point prompts introduce more ambiguity compared to box prompts, which are geometrically more definite, providing no geometric boundaries for the frozen decoder to guide its prediction, resulting in significant performance degradation when no parameter updates are available.}
    
    
     Across the four datasets, Dice scores increase from 10.53 to 83.78 (BRISC), 1.64 to 72.58 (LungSegDB), 8.04 to 76.32 (BUSI), and 5.68 to 89.05 (FLARE22). \textcolor{black}{These gains show that even approximate, automatically estimated spatial priors provide MedSAM with the type of geometric context it was trained to expect from box prompts. The Box Predictor acts as a plug-in module that restores MedSAM's intended operating conditions, allowing the frozen model to recover high accuracy with minimal added complexity.} \textcolor{black}{This recovery, however, should be interpreted as reinstating a familiar prompt geometry for MedSAM rather than as a general solution to point-prompt ambiguity. A measurable gap still separates predicted boxes from oracle-quality boxes in Table~\ref{tab:ablation}, particularly for the low-contrast and fine-grained structures in BUSI and BRISC. Consistent with this, the failure cases in Figure~\ref{fig:worse_cases} show that inaccurate predicted boxes can still degrade segmentation relative to the point-only baseline.}

    \textcolor{black}{Table~\ref{tab:ablation} also reports the inference time and trainable parameter count for each configuration. All timing measurements were obtained on a single NVIDIA RTX 4090 GPU. Across all prompt combinations, the addition of the Box Predictor introduces a negligible inference overhead, with all configurations completing in approximately 61--63 ms per image. Furthermore, the Box Predictor achieves these improvements with only 1.6M trainable parameters, indicating that segmentation performance gains can be obtained at minimal computational cost.}

    Interestingly, combining the predicted box with the input point does not further improve performance and sometimes decreases it. This suggests that when MedSAM is frozen, the predicted box alone provides sufficient guidance, and adding the original point may introduce conflicting information.
    
    When MedSAM is fine-tuned, the benefit of predicted boxes remains evident. In this setting, the prompt encoder and mask decoder adapt to the dataset, yet using predicted boxes continues to provide meaningful improvements. The effect is particularly clear in datasets with high anatomical variability, such as BUSI (87.71 $\to$ 89.04, +1.33 points) and FLARE22 (91.69 $\to$ 93.14, +1.45 points), showing that the predicted boxes contain spatial cues that align with the underlying object structure.
    
    For datasets where point-based fine-tuning already achieves high performance, such as LungSegDB (97.88), the inclusion of predicted boxes leads to a modest but consistent improvement, with Dice scores increasing to 98.16 when using either the box alone or the box–point combination. This indicates that, even in scenarios where the baseline performance is strong, the Box Predictor can still contribute useful spatial guidance without destabilizing the model. Across all configurations, fine-tuned MedSAM remains stable, and incorporating predicted boxes either improves performance or matches the best alternative.
    
     \textcolor{black}{To establish an upper bound on the achievable performance, we also evaluated MedSAM using ground truth bounding boxes as prompts (Perfect Box). In the frozen setting, Perfect Box achieves Dice scores of 0.8950, 0.9566, 0.9259, and 0.9392 on BRISC, LungSegDB, BUSI, and FLARE22 respectively, while in the trainable setting these increase further to 0.9320, 0.9855, 0.9576, and 0.9603. The gap between our predicted box results and the Perfect Box upper bound varies across datasets. The largest gap is observed on BUSI, which is expected given that ultrasound is the lowest contrast modality in our evaluation, where poorly defined lesion boundaries make precise localization from local embedding patches inherently more difficult. BRISC shows the second largest gap, which can be attributed to the complex and heterogeneous texture of brain tumor images that makes accurate boundary estimation challenging.}  \textcolor{black}{These Perfect Box gaps highlight the remaining localization headroom of the proposed approach: improving box quality could further increase Dice from 0.8904 to 0.9576 on BUSI and from 0.8793 to 0.9320 on BRISC. Thus, while predicted boxes provide useful box-style priors for MedSAM, box localization quality still clearly matters.}

	\subsubsection{\textcolor{black}{Box Predictor Regression Performance}}

\begin{figure*}
	\centering
	\begin{tabular}{cc}
		\includegraphics[width=0.48\linewidth]{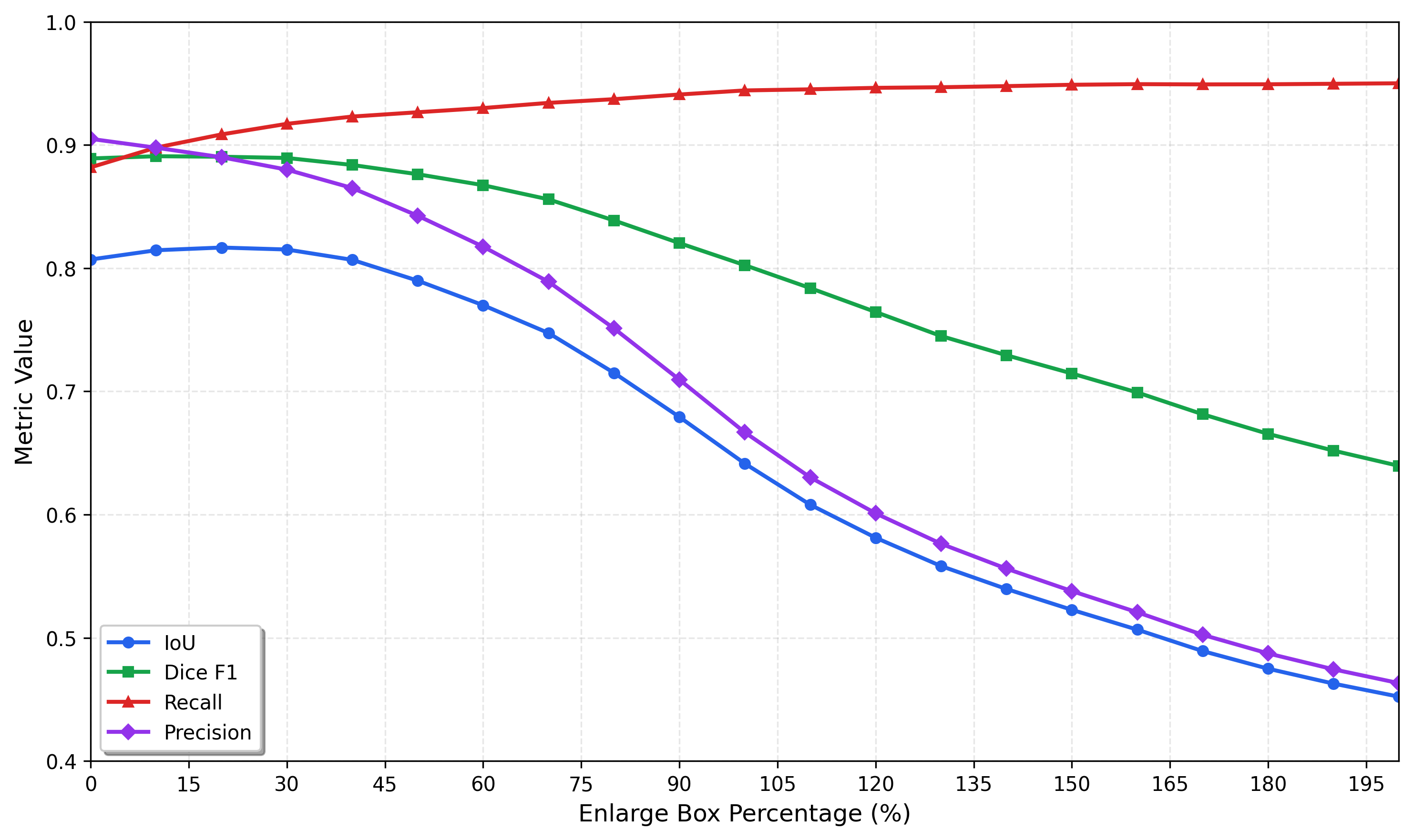} & 
		\includegraphics[width=0.48\linewidth]{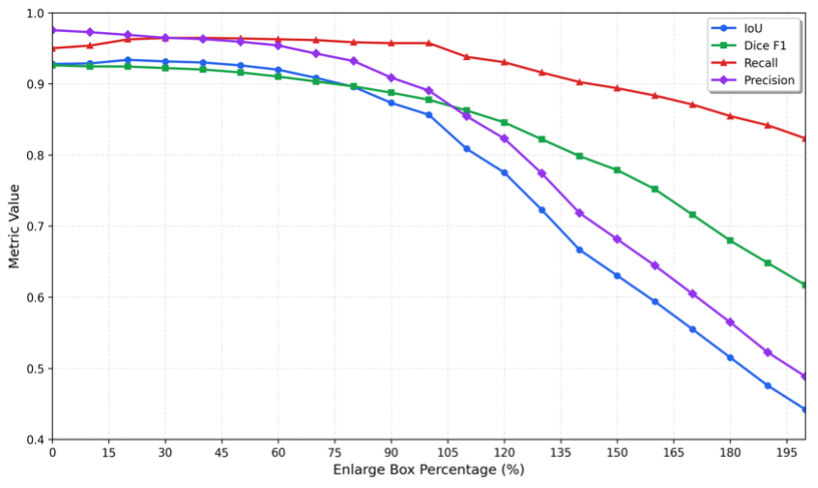} \\
		(a) BUSI (Ultrasound) & (b) FLARE22 (CT) \\[6pt]
		\includegraphics[width=0.48\linewidth]{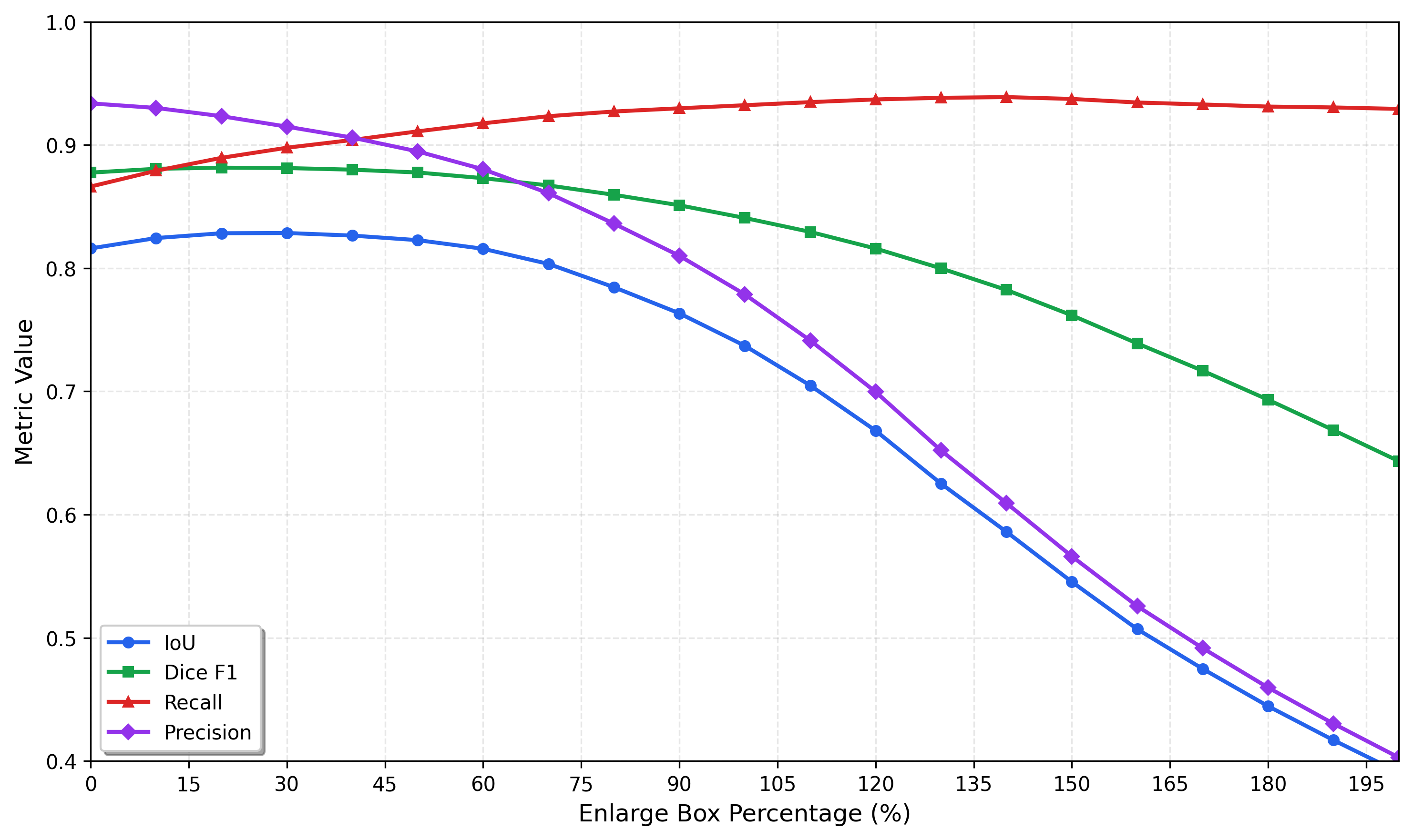} & 
		\includegraphics[width=0.48\linewidth]{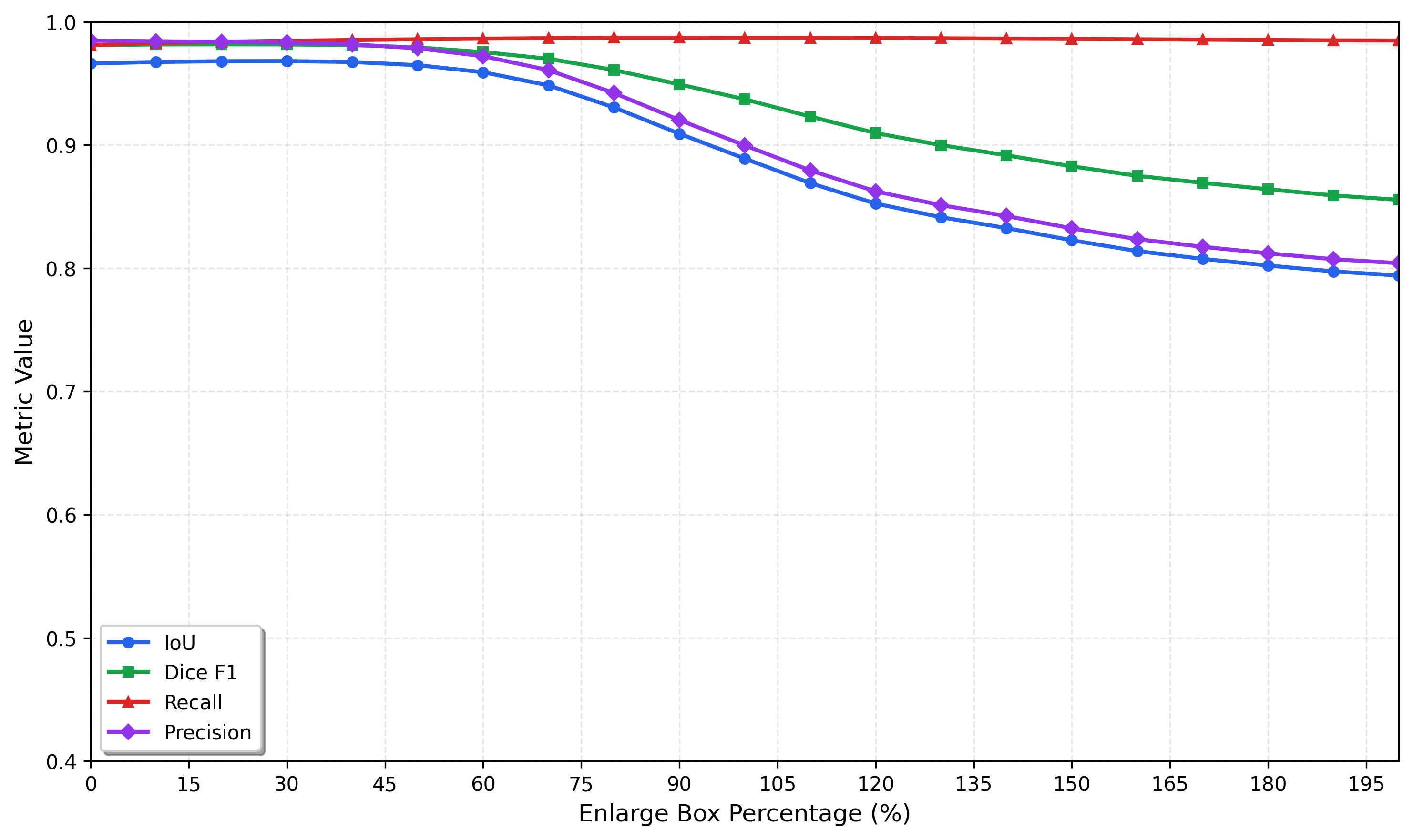} \\
		(c) BRISC (MRI) & (d) LungSegDB (CT)
	\end{tabular}
	\caption{Performance metrics vs. box enlargement percentage across four datasets. The x-axis represents the percentage of uniform enlargement applied to the predicted bounding boxes. The curves illustrate the trade-off between Recall (red) and Precision (purple) and their impact on Dice (green) and IoU (blue).}
	\label{fig:box_enlarge}
\end{figure*}

\textcolor{black}{Table~\ref{tab:box_metrics} reports the bounding box regression performance of the Box Predictor across the four datasets. The results are consistent with the characteristics of each imaging modality: LungSegDB achieves the highest scores (Dice: 0.8766, IoU: 0.7906), reflecting the well-defined and spatially stable nature of lung boundaries in CT imaging. BRISC follows with a Dice of 0.7916, while BUSI and FLARE22 present more challenging cases, achieving Dice scores of 0.7469 and 0.7559 respectively, owing to irregular lesion shapes, low contrast, and high anatomical variability.}\textcolor{black}{The box-regression results show that the predicted boxes are sufficiently accurate to provide useful spatial guidance, but the Perfect Box upper-bound results indicate that further improvements in localization quality could still translate into better segmentation performance.}

\begin{table}[h]
	\centering
	\caption{\textcolor{black}{Bounding box regression performance of the Box Predictor across four datasets, evaluated on the same test splits and perturbed-point protocol as Tables~\ref{tab:comparison} and~\ref{tab:ablation}.}}
	
	\label{tab:box_metrics}
	\renewcommand{\arraystretch}{1.3}
	\begin{tabular}{lcccc}
		\hline
		\textcolor{black}{\textbf{Dataset}} &
		\textcolor{black}{\textbf{IoU}} &
		\textcolor{black}{\textbf{Dice}} &
		\textcolor{black}{\textbf{Recall}} &
		\textcolor{black}{\textbf{Precision}} \\ \hline
		\textcolor{black}{BUSI} &
		\textcolor{black}{0.6117} &
		\textcolor{black}{0.7469} &
		\textcolor{black}{0.7880} &
		\textcolor{black}{0.7844} \\
		\textcolor{black}{BRISC} &
		\textcolor{black}{0.6743} &
		\textcolor{black}{0.7916} &
		\textcolor{black}{0.8640} &
		\textcolor{black}{0.7840} \\
		\textcolor{black}{FLARE22} &
		\textcolor{black}{0.6314} &
		\textcolor{black}{0.7559} &
		\textcolor{black}{0.7913} &
		\textcolor{black}{0.7867} \\
		\textcolor{black}{LungSegDB} &
		\textcolor{black}{0.7906} &
		\textcolor{black}{0.8766} &
		\textcolor{black}{0.9050} &
		\textcolor{black}{0.8653} \\ \hline
	\end{tabular}
\end{table}

\begin{figure}[t]
    \centering
    \includegraphics[width=\columnwidth]{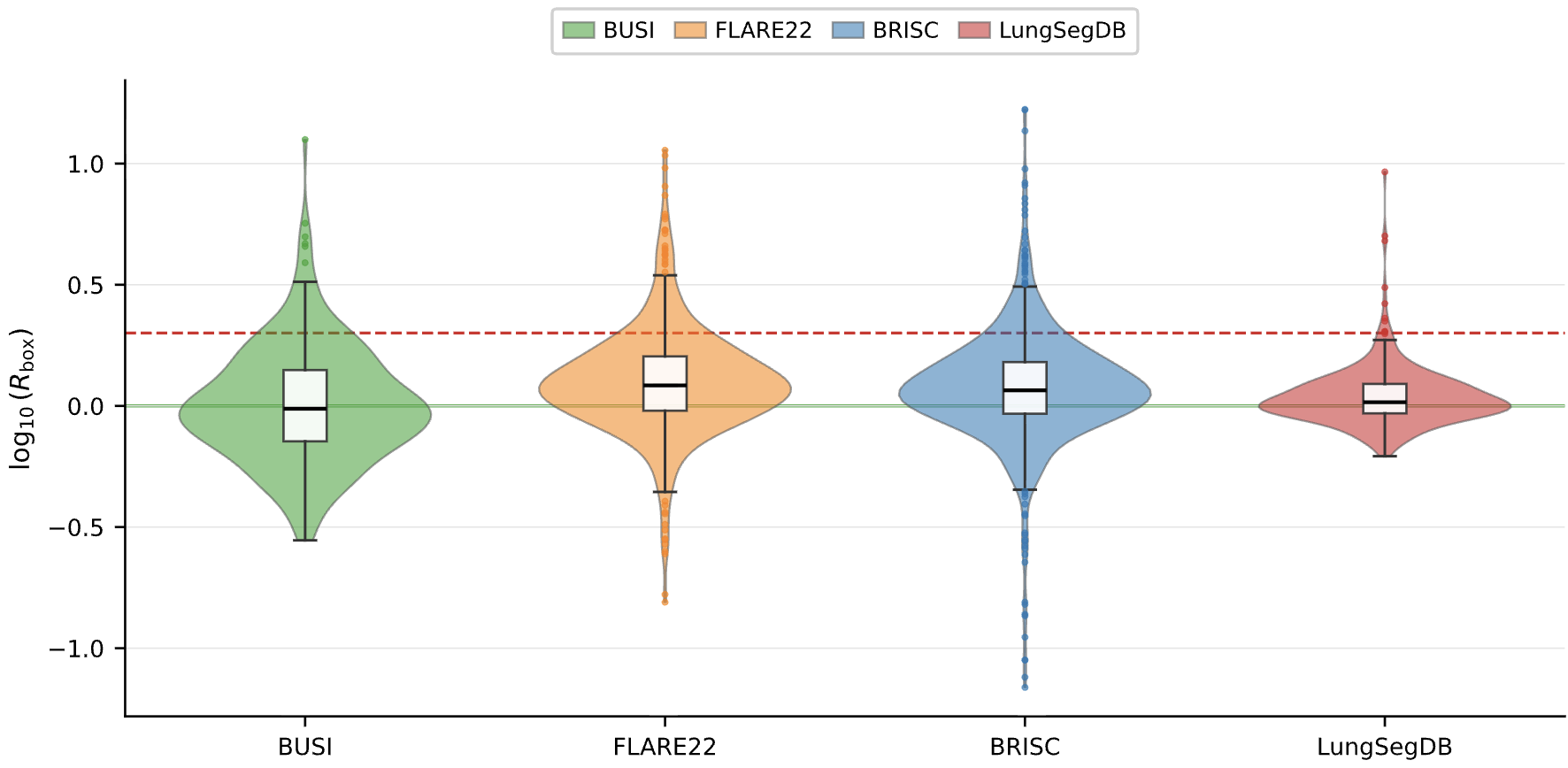}
    \caption{
    \textcolor{black}{Box area-ratio analysis over the test set. 
    $R_{\mathrm{box}} = |B_{\mathrm{pred}}|/|B_{\mathrm{gt}}|$ measures the relative area of the predicted bounding box compared with the ground-truth bounding box. 
    The distribution is shown on a $\log_{10}$ scale, where zero indicates equal predicted and ground-truth box area. 
    The red dashed line marks the severe over-estimation threshold ($R_{\mathrm{box}}>2$).
    }}
    \label{fig:area_ratio_analysis}
\end{figure}

\begin{table}[t]
	\centering
	\caption{\textcolor{black}{Severe box over-estimation stratified by dataset-specific object-size quartiles. Each cell reports the percentage and number of cases with $R_{\mathrm{box}}>2$.}}
	\label{tab:severe_overbox_by_quartile_size}
	\renewcommand{\arraystretch}{1.2}
	\setlength{\tabcolsep}{10pt}
	\resizebox{\columnwidth}{!}{%
		\begin{tabular}{lccc}
			\toprule
			\textcolor{black}{\textbf{Dataset}} &
			\textcolor{black}{\textbf{Small}} &
			\textcolor{black}{\textbf{Medium}} &
			\textcolor{black}{\textbf{Large}} \\
			\midrule
			\textcolor{black}{BUSI} &
			\textcolor{black}{34.9\% (22/63)} &
			\textcolor{black}{0.8\% (1/125)} &
			\textcolor{black}{0.0\% (0/63)} \\
			\textcolor{black}{FLARE22} &
			\textcolor{black}{36.5\% (57/156)} &
			\textcolor{black}{5.2\% (16/310)} &
			\textcolor{black}{0.0\% (0/155)} \\
			\textcolor{black}{BRISC} &
			\textcolor{black}{38.1\% (82/215)} &
			\textcolor{black}{1.9\% (8/430)} &
			\textcolor{black}{0.0\% (0/215)} \\
			\textcolor{black}{LungSegDB} &
			\textcolor{black}{10.5\% (9/86)} &
			\textcolor{black}{0.6\% (1/171)} &
			\textcolor{black}{0.0\% (0/86)} \\
			\midrule
			\textcolor{black}{Overall} &
			\textcolor{black}{32.7\% (170/520)} &
			\textcolor{black}{2.5\% (26/1036)} &
			\textcolor{black}{0.0\% (0/519)} \\
			\bottomrule
		\end{tabular}%
	}
\end{table}

\textcolor{black}{Following the area-ratio failure analysis style used in recent SAM-family segmentation studies~\cite{47}, we define the box-area ratio as $R_{\mathrm{box}}=|B_{\mathrm{pred}}|/|B_{\mathrm{gt}}|$ and the mask-area ratio as $R_{\mathrm{mask}}=|M_{\mathrm{pred}}|/|M_{\mathrm{gt}}|$, where values larger than one indicate over-estimation and $R>2$ is treated as severe over-estimation. Figure~\ref{fig:area_ratio_analysis} shows that most predicted boxes are concentrated around $R_{\mathrm{box}}=1$, suggesting that the Box Predictor does not have a strong global over-estimation bias. However, the upper tails crossing $R_{\mathrm{box}}>2$ indicate that severe over-estimation still occurs in a subset of cases, especially in BUSI, FLARE22, and BRISC. To quantify where these failures occur, we stratify severe box over-estimation by object size in Table~\ref{tab:severe_overbox_by_quartile_size}. Object-size bins are defined separately within each dataset using ground-truth mask-area quartiles: Small corresponds to area $\leq Q1$, Medium to $Q1<$ area $\leq Q3$, and Large to area $>Q3$.}

\textcolor{black}{Table~\ref{tab:severe_overbox_by_quartile_size} shows that severe box over-estimation is primarily a small-object issue. Overall, 32.7\% of small objects show $R_{\mathrm{box}}>2$, compared with only 2.5\% of medium objects and 0.0\% of large objects. The corresponding mask-area analysis shows that these box-level errors are often attenuated by the MedSAM decoder: severe mask over-segmentation, defined as $R_{\mathrm{mask}}>2$, occurs in only 6.5\% of small objects, 0.3\% of medium objects, and 0.0\% of large objects. Thus, the main limitation is small-object box localization, while many box-size errors do not become severe final-mask over-segmentation.}

\color{black}
\subsubsection{Effect of Local Patch Size}

\begin{table}
	\centering
	\caption{\color{black}Effect of embedding patch size on segmentation performance on the BRISC dataset using the predicted box prompt with an input point. Best results are shown in bold.}
	\label{tab:patch_size}
	\renewcommand{\arraystretch}{1.5}
	\begin{tabular}{c c c c c c}
		\hline  	 		
		\color{black}\textbf{Patch Size} & \color{black}\textbf{Recall} & \color{black}\textbf{Precision} & \color{black}\textbf{IoU}    & \color{black}\textbf{Dice} & \multicolumn{1}{c}{\shortstack{\textbf{\textcolor{black}{Trainable}}\\\textbf{\textcolor{black}{Params (M)}}}} \\ \hline
		\color{black}$3\times3$ & \color{black}\textbf{0.9004} & \color{black}0.8848          & \color{black}0.8019 & \color{black}0.8779 & \textcolor{black}{0.6} \\
		\color{black}$5\times5$ & \color{black}0.8992 &\color{black} 0.8903            & \color{black}\textbf{0.8059} & \color{black}\textbf{0.8815}    & \textcolor{black}{1.6} \\
		\color{black}$7\times7$ & \color{black}0.8884 &\color{black} 0.8953 & \color{black}0.8009 & \color{black}0.8763 & \textcolor{black}{3.2} \\
		\color{black}$9\times9$ & \color{black}0.8774 & \color{black}\textbf{0.9081}          & \color{black}0.8034 & \color{black}0.8789 & \textcolor{black}{5.3} \\ \hline
	\end{tabular}
\end{table}

    The Box Predictor module utilizes a local patch extracted from the image embedding to predict the bounding box parameters. The size of this patch determines the amount of contextual information available to the MLP. A patch that is too small may fail to capture sufficient local context, while an excessively large patch might introduce irrelevant background noise or global features that distract the lightweight predictor.
    
    \textcolor{black}{To justify the choice of a $5\times5$ patch size, we conducted an ablation study on the BRISC dataset by varying the patch size from $3\times3$ to $9\times9$. BRISC was selected as the representative dataset for this ablation because its heterogeneous brain tumor boundaries and complex MRI texture patterns make it a particularly demanding setting, where the effect of increasing contextual patch size and MLP capacity is most likely to be pronounced. The results are summarized in Table \ref{tab:patch_size}. As observed, the $5\times5$ patch size yields the highest IoU ($0.8059$) and Dice score ($0.8815$). While the $3\times3$ patch achieves the highest Recall ($0.9004$), it suffers from reduced Precision ($0.8848$) and lower Dice ($0.8779$), reflecting that insufficient contextual information results in less accurate boundary estimation. On the other hand, the $9\times9$ patch achieves the highest Precision ($0.9081$) but experiences a notable Recall decline ($0.8774$), suggesting that larger patches introduce irrelevant background features that cause overly conservative predictions. Beyond segmentation quality, increasing patch size also leads to a proportional growth in trainable parameters, from 0.6M ($3\times3$) to 5.3M ($9\times9$), without yielding corresponding performance gains. The $5\times5$ configuration, with 1.6M parameters, strikes the optimal balance between contextual richness, parameter efficiency, and segmentation accuracy, and was therefore adopted for all experiments across all four datasets.}
    
			\color{black}

	\subsubsection{Effect of Bounding Box Size Adjustment}

    The BoxPredictor generates bounding boxes based on local image patches, which may not always perfectly match the true object boundaries. To analyze the robustness of our method to box size variations and identify optimal expansion factors, we conducted experiments with systematic box enlargement across all four datasets (BUSI, FLARE22, BRISC, and LungSegDB). For each predicted bounding box, we applied uniform enlargement percentages ranging from 0\% to 200\% in 10\% increments. Figure \ref{fig:box_enlarge} illustrates the impact of this adjustment on segmentation performance.
      \begin{figure}[h]
    	\centering
    	
    	\centering
    	\includegraphics[width=0.9\linewidth]{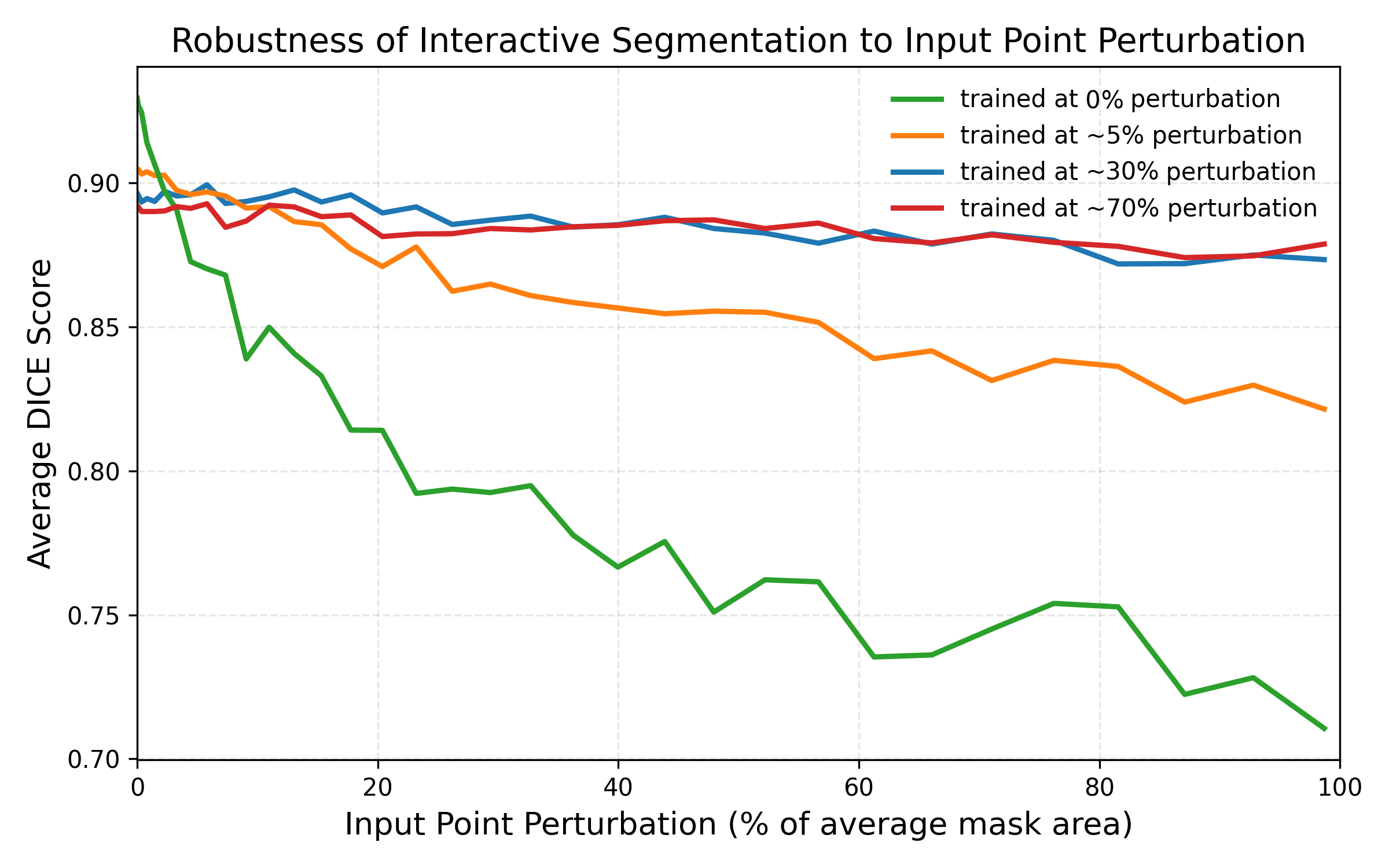}
    	\caption{Effect of input point perturbation on segmentation accuracy (BUSI dataset).}
    	\label{fig:perturbation}
    \end{figure}
    \textbf{BUSI (Ultrasound) \& BRISC (MRI):} As seen in Figure \ref{fig:box_enlarge}(a) and Figure \ref{fig:box_enlarge}(c), these datasets exhibit a similar trend characteristic of modalities with softer or more complex boundaries. Both show a slight performance peak in the 10-20\% enlargement range. For instance, in BRISC, Recall (red curve) steadily increases from $\sim$0.87 to over 0.90 as the box expands, indicating that the initial tight boxes might miss subtle tumor boundaries. However, Precision (purple curve) drops sharply after 30\%, leading to a decline in the overall Dice score. This suggests that a moderate enlargement (10-20\%) is beneficial for these modalities to ensure complete object coverage without introducing excessive background.

   \begin{table*}[h]
   	\centering
   	\caption{Performance comparison between the proposed MedSAM + BoxPredictor and the baseline MedSAM under specific perturbation scenarios. ``Edge Point'' represents an extreme stress test where the input is placed exactly on the mask boundary.}
   	\label{tab:edge_perturbation}

   	\renewcommand{\arraystretch}{1.2}
   	
   	\setlength{\tabcolsep}{8pt}
   	
   	\begin{tabular}{llcccc}
   		\toprule
   		\textbf{Scenario} & \textbf{Model} & \textbf{Recall} & \textbf{Precision} & \textbf{IoU} & \textbf{Dice} \\
   		\midrule
   		
   		\multirow{2}{*}{0 Perturbation} 
   		& MedSAM & 0.8900 & \textbf{0.9474} & 0.8459 & 0.9142 \\
   		& MedSAM + BoxPredictor & \textbf{0.9238} & 0.9406 & \textbf{0.8716} & \textbf{0.9291} \\
   		\addlinespace 
   		
   		\multirow{2}{*}{30\% Area Prompt} 
   		& MedSAM & 0.8945 & \textbf{0.8843} & 0.7936 & 0.8771 \\
   		& MedSAM + BoxPredictor & \textbf{0.9176} & 0.8814 & \textbf{0.8124} & \textbf{0.8904} \\
   		\addlinespace
   		
   		\multirow{2}{*}{70\% Area Prompt} 
   		& MedSAM & \textbf{0.8951} & 0.8607 & 0.7720 & 0.8602 \\
   		& MedSAM + BoxPredictor & 0.8720 & \textbf{0.9163} & \textbf{0.8026} & \textbf{0.8830} \\
   		\addlinespace
   		
   		\multirow{2}{*}{Edge Point} 
   		& MedSAM & \textbf{0.8718} & 0.8500 & 0.7551 & 0.8435 \\
   		& MedSAM + BoxPredictor & 0.8408 & \textbf{0.8903} & \textbf{0.7672} & \textbf{0.8516} \\
   		
   		\bottomrule
   	\end{tabular}
   \end{table*}
   \begin{figure*}
   	\centering
   	
   	\centering
   	\includegraphics[width=0.9\linewidth]{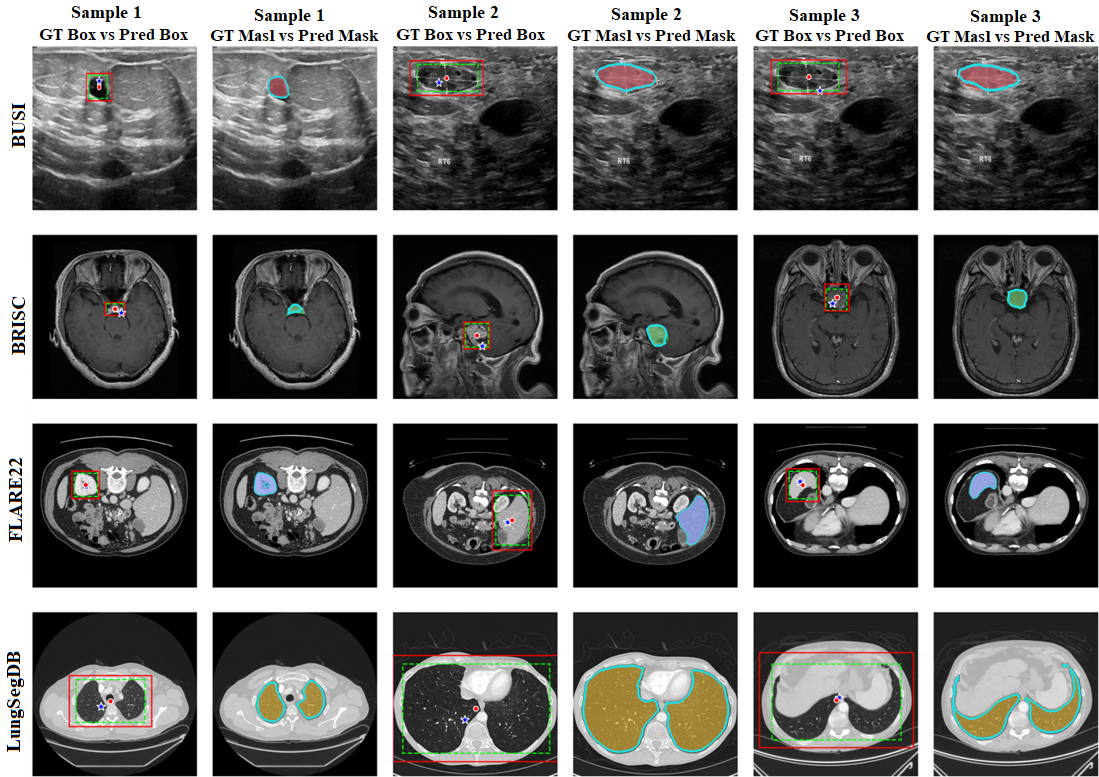}
   	\caption{Qualitative segmentation results across four datasets. Left image: Ground truth masks (colored overlays) compared with predicted masks (cyan boundaries). Right image: Bounding box predictions showing ground truth boxes (green, dashed), predicted boxes (red, solid), user point prompts (blue stars), and predicted box centers (red dots).}
   	\label{fig:qualitative}
   \end{figure*}
    
    \textbf{FLARE22 \& LungSegDB (CT):} The results for the CT datasets, shown in Figure \ref{fig:box_enlarge}(b) and Figure \ref{fig:box_enlarge}(d), demonstrate higher initial stability. For LungSegDB, the Recall is exceptionally high ($\sim$0.99) and remains flat, meaning the BoxPredictor captures the lung region almost perfectly from the start. Any enlargement immediately causes a gradual drop in Precision and Dice, indicating that the predicted boxes are already optimal. FLARE22 shows a stable plateau between 0\% and 30\%, where metrics remain high before degrading.

    The contrasting patterns reveal a modality-specific characteristic. Ultrasound and MRI segmentation (BUSI, BRISC) benefit from slight box enlargement to compensate for ambiguous boundaries and irregular lesion shapes. In contrast, CT segmentation (FLARE22, LungSegDB) achieves near-optimal performance with the initial tighter boxes due to well-defined anatomical contrast. Despite these differences, all datasets maintain robust performance within a safe margin (0-20\% enlargement), confirming that the BoxPredictor provides reliable localization that does not require extensive manual tuning.

\begin{figure*}[t]
    \centering
    \includegraphics[width=0.95\textwidth]{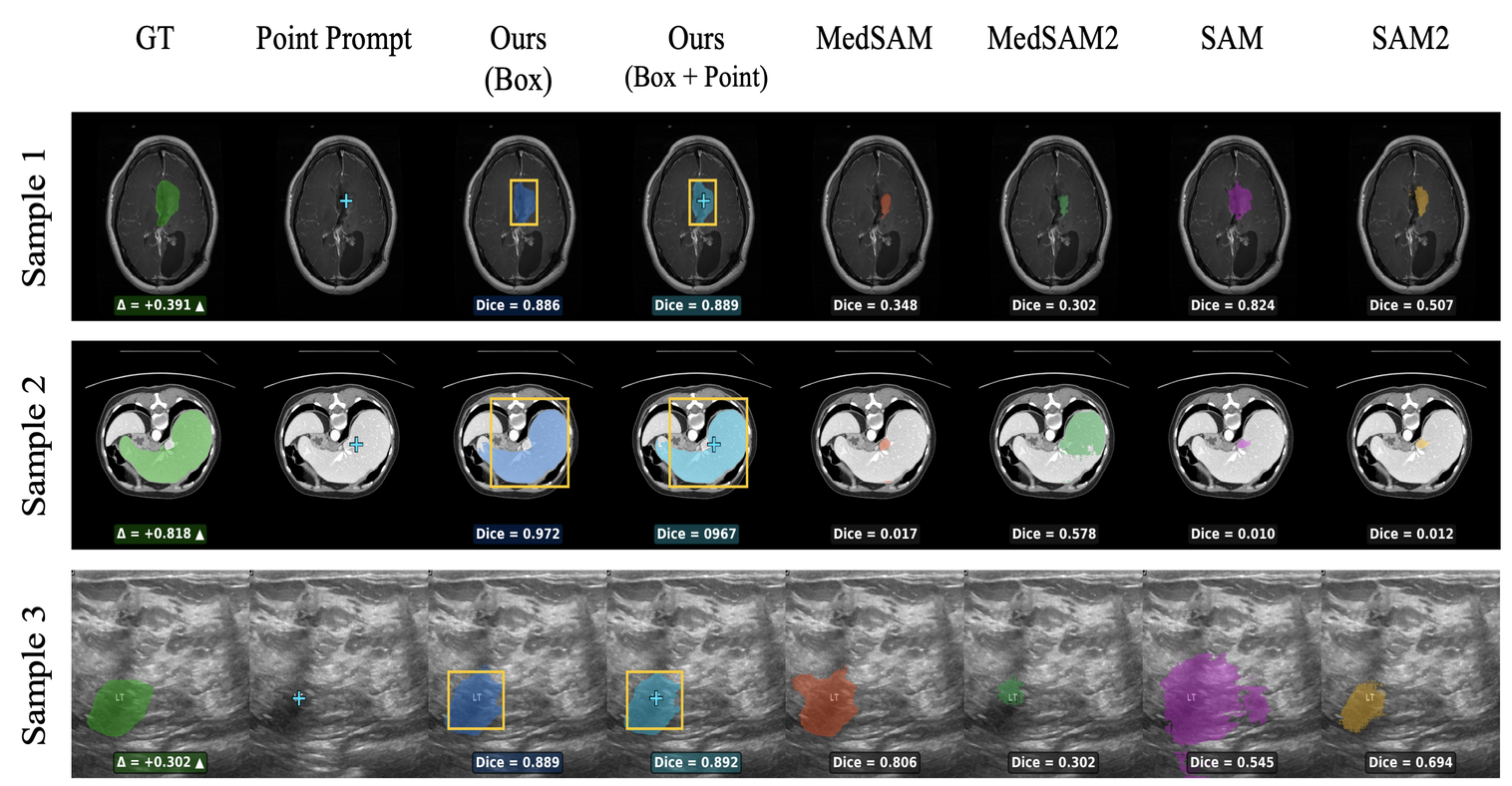}
    \caption{
    \textcolor{black}{Representative qualitative examples where the proposed Box Predictor improves segmentation accuracy.
    ``Ours (Box)'' uses the predicted bounding box as the segmentation prompt, while ``Ours (Box + Point)'' uses both the predicted box and the original point prompt. 
    The MedSAM, MedSAM2, SAM, and SAM2 baselines are evaluated using the point prompt shown in the Point Prompt column. 
    Dice scores are reported below each predicted mask. 
    Sample 1 is from BRISC, Sample 2 is from FLARE22, and Sample 3 is from BUSI.
    }}
    \label{fig:better_cases}
\end{figure*}

    \subsubsection{Effect of Input Point Perturbation}
    \label{sec:perturbation}

    In real-world interactive segmentation, user clicks are rarely perfect geometric centroids. Users often click near object boundaries or vaguely within the target region. To ensure our Box Predictor is robust to such variances, we analyzed the impact of training with different levels of spatial perturbation. Unlike previous approaches that use fixed pixel distances, we define perturbation as a percentage of the average mask area to maintain scale invariance across different datasets.
    
    Figure \ref{fig:perturbation} illustrates the robustness of the model when trained with varying degrees of point perturbation (~0\%, ~5\%, ~30\%, and ~70\%) and evaluated across a wide range of inference-time perturbations (0\% to 100\% of the mask size).

    As observed in Figure \ref{fig:perturbation}, models trained with minimal perturbation (0\% and 5\%, represented by green and orange lines) achieve high Dice scores when the input point is perfectly centered. However, their performance degrades significantly as the test point moves away from the center. For instance, the model trained with 0\% perturbation sees a sharp drop in Dice score from $\sim$0.93 to below 0.75 when the point is perturbed by 50\% or more. This indicates that without perturbation during training, the model overfits to the centroid and fails to generalize to off-center clicks (e.g., clicks near the mask edge).
    
    In contrast, models trained with higher perturbation levels (~30\% and ~70\%, represented by blue and red lines) demonstrate exceptional stability. The curve for the 30\% model is nearly flat, maintaining a Dice score between 0.88 and 0.90 regardless of whether the input point is at the center or the edge of the mask. Although these models sacrifice a marginal amount of peak performance at the absolute zero-perturbation point compared to the 0\% model, they offer superior reliability for practical usage.
    
    Consequently, we selected $\sim$30\% perturbation for our final training configuration. This setting provides the optimal trade-off, ensuring the model remains robust to user errors and boundary clicks without the excessive variance potentially introduced by extreme perturbation levels.

    To further validate this robustness, we evaluated the model under discrete "stress test" scenarios, as shown in Table \ref{tab:edge_perturbation}. We specifically included the "Edge Point" scenario, where the user click is placed exactly on the boundary of the object. While this case is clinically rare (users almost always click somewhere within the central region of the lesion) it serves as an important worst-case benchmark. As expected, performance metrics drop in this extreme scenario compared to the centered (0 perturbation) case. However, the results remain functional, with the Dice score staying above 0.84. The Box Predictor remains beneficial in this stress-test setting, achieving higher Dice and IoU than the baseline MedSAM. This indicates that even when the user provides a suboptimal click on the very edge of the object, the predicted bounding box helps the model "recover" and cover a larger portion of the target region than the point-only baseline would allow.

\begin{figure*}[t]
    \centering
    \includegraphics[width=0.95\textwidth]{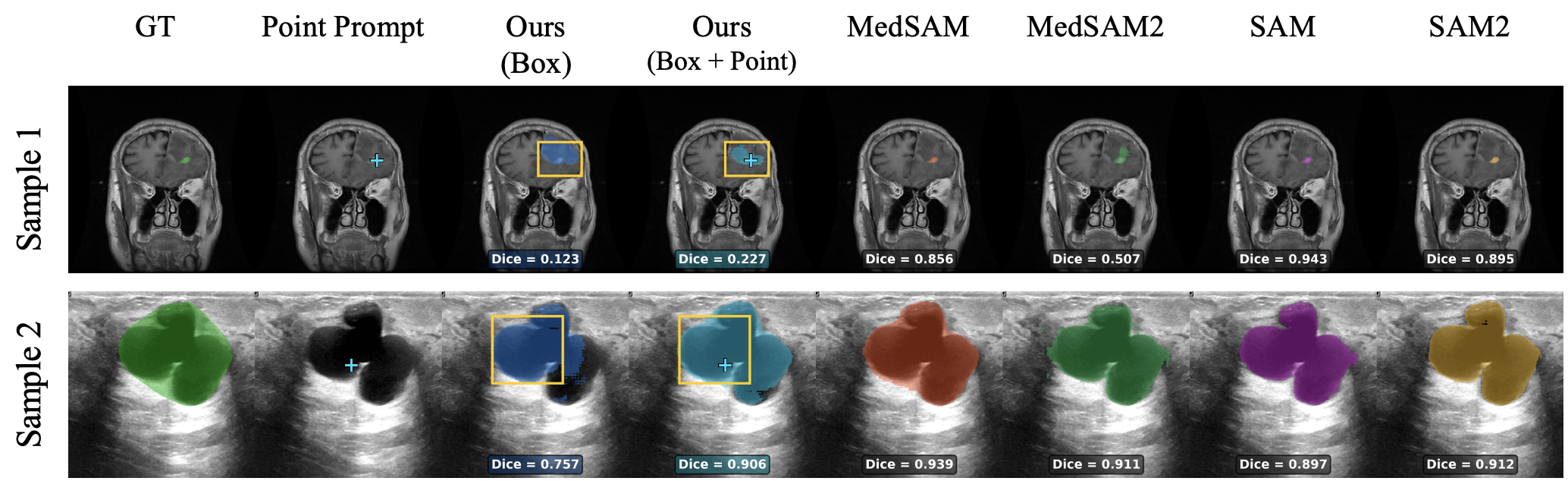}
    \caption{
    \textcolor{black}{Representative limitation cases of the proposed Box Predictor. 
    ``Ours (Box)'' uses only the predicted bounding box as the segmentation prompt, while ``Ours (Box + Point)'' uses both the predicted box and the original point prompt. 
    The MedSAM, MedSAM2, SAM, and SAM2 baselines are evaluated using the point prompt shown in the Point Prompt column. 
    Dice scores are reported below each predicted mask. 
    Sample 1 is from BRISC and Sample 2 is from BUSI.
    }}
    \label{fig:worse_cases}
\end{figure*}

\textcolor{black}{One aspect of this protocol that requires clarification is the resampling step. When a perturbed point falls outside the target mask, it is replaced by a point sampled uniformly from within the valid mask region. Therefore, the effective perturbation distribution can be more conservative than the nominal 30\% setting, especially for thin, irregular, or non-convex masks. Table~\ref{tab:resampling} reports the fraction of samples requiring resampling for each dataset. For BUSI, BRISC, and FLARE22, the resampling rates are low (3.3--9.2\%), suggesting that the 30\%-area perturbation largely behaves as intended. LungSegDB is an exception, with resampling exceeding 50\% in both splits. This is likely due to the bilateral and non-convex geometry of lung masks, where centroid-based perturbations frequently fall outside the foreground region. Therefore, the LungSegDB perturbation results should be interpreted with this protocol characteristic in mind. Nevertheless, LungSegDB remains useful for comparison because the same prompting protocol is applied consistently across all interactive methods, and the dataset exhibits near-ceiling segmentation performance with small differences among prompt-based approaches.}

\begin{table}[h]
	\centering
	\caption{\textcolor{black}{Fraction of samples requiring point resampling (count/total (\%)).}}
	\label{tab:resampling}
	\renewcommand{\arraystretch}{1.3}
	\resizebox{\columnwidth}{!}{%
		\begin{tabular}{lcc}
			\hline
			\textcolor{black}{\textbf{Dataset}} &
			\textcolor{black}{\textbf{Train}} &
			\textcolor{black}{\textbf{Test}} \\
			\hline
			\textcolor{black}{BUSI} &
			\textcolor{black}{{\footnotesize 32/678\ \ (4.7\%)}} &
			\textcolor{black}{{\footnotesize 9/251\ \ (3.6\%)}} \\
			\textcolor{black}{BRISC} &
			\textcolor{black}{{\footnotesize 215/3933\ (5.5\%)}} &
			\textcolor{black}{{\footnotesize 28/860\ (3.3\%)}} \\
			\textcolor{black}{FLARE22} &
			\textcolor{black}{{\footnotesize 276/3000\ (9.2\%)}} &
			\textcolor{black}{{\footnotesize 41/621\ (6.6\%)}} \\
			\textcolor{black}{LungSegDB} &
			\textcolor{black}{{\footnotesize 764/1371\ (55.7\%)}} &
			\textcolor{black}{{\footnotesize 180/343\ (52.5\%)}} \\
			\hline
		\end{tabular}%
	}
\end{table}

\subsection{Qualitative Evaluation and Limitation Analysis}

As previously noted, the bounding boxes predicted by the Box Predictor are approximate and may slightly deviate from the ground truth. However, this deviation is generally minor, and qualitative inspection demonstrates that the results are satisfactory.

    The strong performance of the Box Predictor, particularly in datasets where target structures vary widely in size, shape, and spatial location, demonstrates that the extracted patch from the image embedding around the input point effectively captures the necessary contextual information. This localized region of the feature space provides sufficient structural cues for inferring the approximate extent and boundaries of the intended object. Moreover, the qualitative results show that the model is capable of correcting suboptimal point prompts. For example, even when the user-provided point lies near the boundary of the object, as observed in sample 3 of the BUSI dataset in Figure~\ref{fig:qualitative}, the Box Predictor successfully shifts the predicted center toward the true central region of the target. Since the model is trained exclusively on such local patches, its ability to estimate accurate bounding boxes across varying scales indicates that it has implicitly learned how high-level semantic features relate to the physical dimensions and spatial distribution of objects. Thus, the predicted box is not generated through simple geometric heuristics but is derived from interpreting texture patterns, density variations, and other visual cues encoded within the embedding.

\textcolor{black}{Figure~\ref{fig:better_cases} presents representative cases where the predicted box substantially improves segmentation quality over the point-only MedSAM baseline. In Sample 1 from BRISC, MedSAM with only the point prompt produces an incomplete tumor mask with a Dice score of 0.348. In contrast, the proposed box-based prompt provides a clearer estimate of the lesion extent, increasing the Dice score to 0.886 with Ours (Box) and 0.889 with Ours (Box + Point). This case shows that the point prompt alone may localize the lesion but fail to describe its full spatial extent, while the predicted box helps recover a more complete tumor region.}

\textcolor{black}{In Sample 2 from FLARE22, the point prompt is highly ambiguous in a dense abdominal CT scene, where adjacent structures can have similar appearance. MedSAM with the point prompt achieves a Dice score of only 0.017, indicating that the point alone is insufficient to guide the decoder to the correct structure. The predicted box resolves much of this ambiguity by explicitly constraining the target region, improving the Dice score to 0.972 with Ours (Box) and 0.967 with Ours (Box + Point). This example highlights the main advantage of the proposed method: the predicted box provides object-level spatial context that is missing from a single point prompt.}

\textcolor{black}{Sample 3 from BUSI illustrates the benefit of the proposed method in ultrasound imaging, where lesions often have weak boundaries and heterogeneous surrounding tissue. MedSAM with the point prompt achieves a Dice score of 0.806, while Ours (Box) and Ours (Box + Point) improve the Dice score to 0.889 and 0.892, respectively. The predicted box helps restrict the decoder to the relevant lesion area and reduces incorrect activation in surrounding tissue. This result suggests that the box prior is especially useful when the image boundary evidence is insufficient for point-only prompting.}

\textcolor{black}{Figure~\ref{fig:worse_cases} shows representative limitation cases. These examples indicate that the predicted box is not uniformly beneficial in all situations, and its effect depends on the relationship between box quality, object size, and boundary characteristics. In Sample 1 from BRISC, the target lesion is very small and the predicted box has only moderate overlap with the ground-truth box, with a box IoU of 53.34\%. Ours (Box) achieves a Dice score of only 0.123, and adding the point prompt improves the result only partially to 0.227, while MedSAM with the point prompt reaches 0.856. This behavior suggests a box-regression limitation for small objects: even a moderate absolute error in the predicted box can produce a large relative overestimation of the object area. When the predicted box covers too much background around a small target, the decoder receives an overly broad spatial prior and may segment irrelevant tissue instead of the lesion.}

\textcolor{black}{Sample 2 from BUSI shows a different limitation mode. In this case, the lesion is large, high-contrast, and visually well localized by the point prompt. MedSAM with the point prompt achieves a Dice score of 0.939. However, the predicted box has low overlap with the ground-truth box, with a box IoU of 43.79\%, indicating inaccurate box regression for this irregular multi-lobed lesion. Ours (Box) decreases to 0.757, showing that the box-only prompt provides a coarse and misaligned spatial prior. When the point prompt is added back, Ours (Box + Point) recovers most of the lost performance and reaches a Dice score of 0.906, showing that the point helps anchor the decoder to the correct lesion despite the imperfect box. Therefore, these failure cases are associated with low predicted-box IoU, but the underlying causes differ: the BRISC case mainly reflects the sensitivity of small objects to box overestimation, whereas the BUSI case reflects inaccurate box prediction for an irregular lesion in a situation where the point prompt alone is already highly effective.}

    \section{Conclusion}

\textcolor{black}{In this study, we addressed the limitations of point-prompt-based MedSAM segmentation by incorporating a Box Predictor that provides additional spatial guidance through automatically estimated bounding boxes.} By generating approximate bounding boxes from a single user click, our approach effectively restores the model's geometric guidance. Experiments across four diverse datasets (BRISC, FLARE22, LungSegDB, and BUSI) demonstrated significant improvements in accuracy and stability. Notably, our ablation studies confirmed the method's robustness to off-center clicks (achieved through percentage-based perturbation training) and its adaptability across different imaging modalities, requiring minimal box adjustment for CT while benefiting from slight enlargement in Ultrasound and MRI. \textcolor{black}{Although the proposed method does not redefine the underlying segmentation paradigm, it provides a practical enhancement to MedSAM by reducing dependency on precise manual prompts while improving robustness across multiple imaging modalities and datasets.}
    
    \textcolor{black}{A current limitation is the reliance on a single input point, which fundamentally cannot specify multiple disconnected regions simultaneously. For structures composed of multiple separate parts or lesions appearing in multiple locations, a single bounding box cannot represent the full target extent, either including large irrelevant regions or leaving parts of the structure unaddressed. \textcolor{black}{In our LungSegDB setting, this issue is less visible because the lung region has distinctive CT contrast and is reliably captured by the local embedding. However, this dataset-specific behavior should not be interpreted as evidence that a single point or a single predicted box is sufficient for bilateral, multi-focal, or disjoint anatomy in general.} Future work will address this by incorporating multi-point inputs, where a Box Predictor generates a box for each point, which can then be fused via union operations or learned deep fusion techniques to construct a comprehensive spatial prior covering the full target structure.}
    
    Finally, we plan to extend the framework by integrating textual prompts via medical vision-language models like MedCLIP. Since the Box Predictor estimates the target's location and size, this spatial information can be converted into automated textual descriptions (e.g., "small lesion in the upper left lung"). Combining spatial, geometric, and semantic prompts will further enhance system flexibility, enabling unified segmentation systems that jointly leverage spatial, geometric, and semantic prompts within a single clinical inference pipeline.

	\bibliographystyle{IEEEtran}
	\bibliography{ref.bib}
    
\end{document}